\documentclass[twoside,11pt]{article}

\usepackage{jmlr2e, float, tabularx, xcolor}

\definecolor{red}{RGB}{230,25,75}
\definecolor{purple}{RGB}{145,30,180}
\definecolor{green}{RGB}{60,180,75}
\definecolor{orange}{RGB}{245,130,48}
\definecolor{white}{RGB}{230,230,230}
\definecolor{blue}{RGB}{0,130,200}
\definecolor{pink}{RGB}{255,0,255}

\ShortHeadings{Aerial Imagery Pixel-level Segmentation}{M.R. Heffels, J. Vanschoren}
\firstpageno{1}

\begin{document}

\title{Aerial Imagery Pixel-level Segmentation}

\author{\name ir. Michael Heffels \email m.r.heffels@student.tue.nl
      \AND
      \name dr. ir. Joaquin Vanschoren \email j.vanschoren@tue.nl \\
      \addr Department of Mathematics and Computer Science\\
      Eindhoven University of Technology\\
      Netherlands}

\editor{}

\maketitle

\begin{abstract}
Aerial imagery can be used for important work on a global scale. Nevertheless, the analysis of this data using neural network architectures lags behind the current state-of-the-art on popular datasets such as PASCAL VOC, CityScapes and Camvid. In this paper we bridge the performance-gap between these popular datasets and aerial imagery data. Little work is done on aerial imagery with state-of-the-art neural network architectures in a multi-class setting. Our experiments concerning data augmentation, normalisation, image size and loss functions give insight into a high performance setup for aerial imagery segmentation datasets. Our work, using the state-of-the-art DeepLabv3+ Xception65 architecture, achieves a mean IOU of 70\% on the DroneDeploy validation set. With this result, we clearly outperform the current publicly available state-of-the-art validation set mIOU (65\%) performance with 5\%. Furthermore, to our knowledge, there is no mIOU benchmark for the test set. Hence, we also propose a new benchmark on the DroneDeploy test set using the best performing DeepLabv3+ Xception65 architecture, with a mIOU score of 52.5\%.
\end{abstract}


\section{Introduction}\label{sec:introduction}
Aerial imagery is analyzed for a broad number of applications. Our work is inspired by the fact that high-resolution aerial imagery analysis helps to address topics such as deforestation \citep{Green1990DeforestationImages}, declining biological diversity \citep{Richards2016Rates20002012}, refugee camp planning and maintenance \citep{Ko2018USAMaintenance}, global poverty \citep{Roser2013GlobalPoverty}, urban planning, precision agriculture, and geographic information system (GIS) updating \citep{Cheng2016AImages}.

In this paper, we strive to bridge the gap between current state-of-the-art image segmentation solutions on non-aerial imagery and the solutions in aerial imagery, by achieving higher mIOU accuracy on more classes. More formally we ask: \textit{How can we improve the current state-of-the-art neural network architectures used for aerial imagery, to obtain new state-of-the-art performance for multi-class pixel level semantic segmentation on aerial imagery?}

To properly answer this question, we first present several key concepts for this area in Section \ref{sec:background}. Next, we discuss the most important related work done in this area in Section \ref{sec:relatedwork}. In Section \ref{sec:model-design} we discuss how we incorporated the techniques found in Section \ref{sec:relatedwork} in a model design strategy. In Section \ref{sec:experiment-design} we elaborate on the choice of the DroneDeploy dataset, as well as how our experiments and adaptations help shape the final model and how this final model performs in practice on the DroneDeploy dataset. Finally in Section \ref{sec:futurework} we present our conclusions and recommend several opportunities for future work in this area.

\section{Background information}\label{sec:background}
In 2020, different challenges of semantic segmentation within Computer Vision are being solved with a mean Intersection-Over-Union (mIOU) of up to 90\% on popular segmentation datasets, such as CityScapes (mIOU 85.1\% on test set \citep{Tao2020HierarchicalSegmentation}), Camvid (mIOU 81.7\% \citep{Zhu2018ImprovingRelaxation}) and PASCAL VOC 2012 (mIOU 90.5\% \citep{Zoph2020RethinkingSelf-training}). However, this performance is not trivial for the same applications on aerial imagery. Moreover, most novel image segmentation neural network architectures and implementations are focused on non-aerial imagery.

With the increasing number of remote sensing devices with high quality spatial resolution, the amount of aerial imagery data also increases. Spatial resolution is the length of one side of a single pixel. For example, an image with a spatial resolution of 1.5 meters (such as the SPOT-7 \citep{AirbusDefenceandSpace2013SPOTIntelligence}) means that a single pixel represents an area on the ground that is 1.5 meters in length and width. Spatial resolutions between 0.41 - 4 m are considered high spatial resolution, whereas spatial resolutions between 30 - 1000 m are considered low spatial resolution \citep{Campbell2011IntroductionEdition}. 

Where in the early days spatial resolutions of 120 meters were the norm, current state-of-the-art imagery satellites have spatial resolutions of $<0.5$ meters \citep{Zhou2003FutureSatellites}. The general trend is that, in recent years, the world is launching more satellites with ever improving spatial resolutions. Smaller spatial resolution means less physical distance between each pixel, which means that these images could be segmented more accurately.

\begin{figure}[H]
 \centering
 \includegraphics[width=1\textwidth]{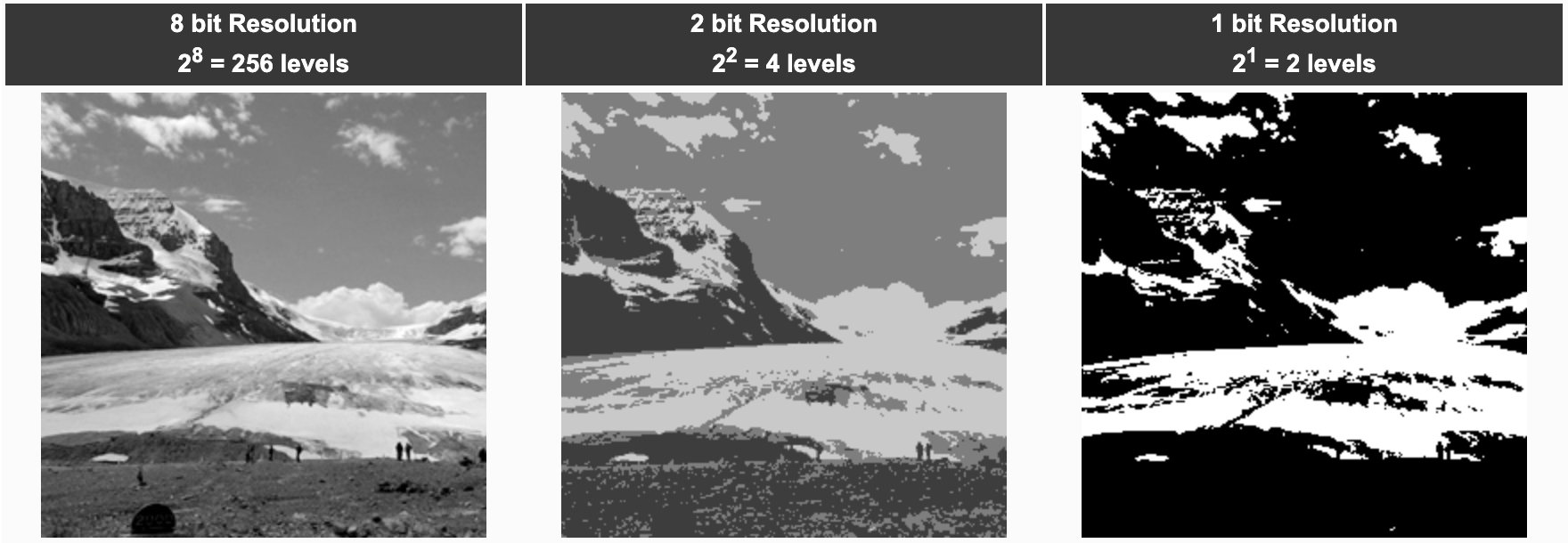}
 \caption{From left to right, 8 bit, 2 bit and 1 bit radiometric resolutions are shown. Adopted from \cite{HumboldtStateUniversity2019GSPResolution}, course introduction to remote sensing}
 \label{fig:radiometric-resolution}
\end{figure}

Noteworthy is that popular regular image datasets (for example CIFAR, ImageNet) typically have a radiometric resolution of 8-bits with RGB values varying between 0 and 255, whereas modern observation satellites have 16-bit RGB values, which means that these are more precisely measured between 0 and $2^{16} = 65536$. Radiometric resolution quantifies the ability to discriminate between slight energy differences. This is stored in a number of bits for each band. For earlier launched satellites 8-bit data was common in remote sensed data, newer sensors (like Landsat 8, launched in 2013) have 16-bit data products. These bits represent the number of different intensities of radiation the sensor is able to distinguish and record, as shown in a Figure from an introductory course \citep{HumboldtStateUniversity2019GSPResolution} in remote sensing in Figure \ref{fig:radiometric-resolution}. Hence, we can safely say that the use of satellite imagery has great potential, as well as challenges for machine learning. 

Traditional aerial imagery analysts are often still manually analysing images and labeling them by hand. In recent years the groundbreaking results of neural networks in computer vision have led to the first neural network developments in the segmentation of images as we will discuss later on in Section \ref{sec:fcn-unet}.

\section{Related work on aerial image segmentation}\label{sec:relatedwork}
This Section discusses fundamental image segmentation techniques on which our baseline experiments are based, different challenges when segmenting aerial images and finally more novel techniques which help obtain pixel-level segmentation on aerial images.
\subsection{Image Segmentation}
\begin{figure}[H]
 \centering
 \includegraphics[width=0.9\textwidth]{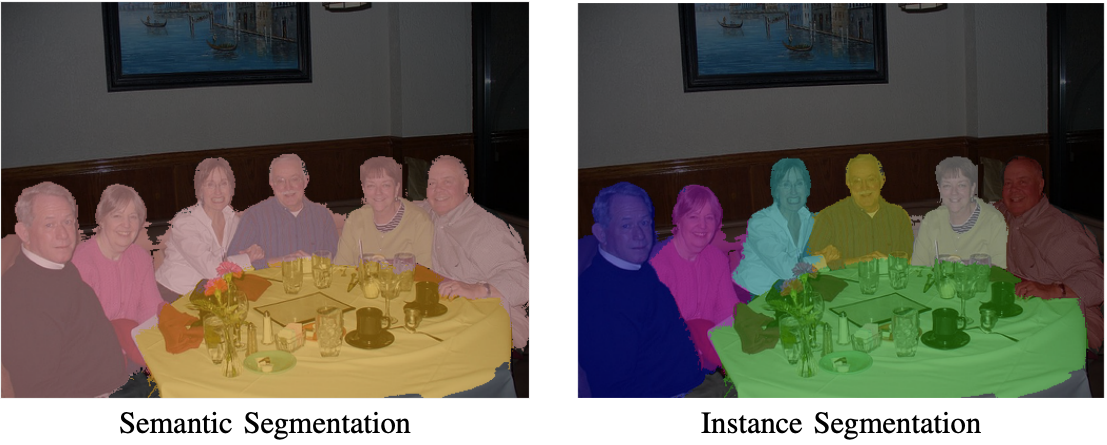}
 \caption{An example of semantic segmentation and instance segmentation from left to right, respectively. (image adopted from \cite{Arnab2018ConditionalPrediction})}
 \label{fig:segmentation-variants}
\end{figure}

Before we discuss the relevant work regarding neural networks and aerial imagery, it is important to note that our proposed solutions focus on some form of segmentation. Namely, semantic or instance segmentation. We show the definition of both types below. These definitions are interpreted from \citep{Arnab2018ConditionalPrediction}:
\begin{definition}[Semantic segmentation]
the process of assigning a label to every pixel in the image, where multiple objects of the same class are treated as the same entity
\end{definition}
\begin{definition}[Instance segmentation]
the process of assigning a label to every pixel in the image, but here multiple objects of the same class are treated as distinct individual objects (or instances). 
\end{definition}
In general, instance segmentation is considered more difficult than semantic segmentation. The difference between both variants is illustrated in Figure \ref{fig:segmentation-variants}.
In short, this means that in order to learn a segmentation mask, we need to classify every pixel in an image. If we color-code each class, this classification at the pixel level will eventually form a segmentation mask such as the masks in Figure \ref{fig:segmentation-variants}. In this paper, we will focus solely on the semantic segmentation task, since we see no clear added value using instance segmentation on aerial imagery at this time. This scope is also specified in our research question.

\subsection{Fundamental image segmentation techniques}\label{sec:fcn-unet}
The well known u-net architecture is a crucial baseline for image segmentation developments. It won the ISBI challenge \citep{Ronneberger2015U-net:Segmentation} for segmentation of neuronal structures in electron microscopic stacks. Furthermore, the authors show that their solution requires a very small number of training examples: it achieved a mean Intersection-Over-Union (mIOU) of $92\%$ on a 'PhC-U373' cells dataset after training on 35 example 512x512 pixel images. This mean Intersection-Over-Union score, shown in equation \ref{eq:miou} is the most used evaluation metric for semantic image segmentation. 

\begin{equation}\label{eq:miou}
    mIOU = \frac{\sum_{i=1}^{C} \frac{A_i \cap B_i}{A_i \cup B_i}}{C}
\end{equation}

In equation \ref{eq:miou}, the intersection ($A_i \cap B_i$) for each class $1 \leq i \leq C$, comprises the pixels found in both the prediction mask $A_i$ and the ground truth mask $B_i$. The union ($A_i \cup B_i$) includes all pixels found in either the prediction mask or the ground truth mask. After computing the IOU for each trainable class, the score is averaged out over the number of classes $C$, with or without taking into account class imbalance. Most of the time the unbalanced version is used, as displayed in equation \ref{eq:miou}.

The u-net architecture is shown below in Figure \ref{fig:u-net_architecture}. It is based on a publication by \citep{Long2014FullySegmentation}, which introduces the Fully Convolutional Network (FCN). A FCN replaces the final dense layers by more convolution layers. Hence, a FCN uses filters to learn representations and make decisions based on local spatial input throughout the entire neural network. This makes sense for our use case, since correctly classifying any pixel in an image strongly depends on local information. One other big advantage of doing this is that the size of input is now flexible. U-nets are very effective for tasks where the output size is similar to the input size, where the output needs to be the same high resolution. This makes them very good for creating segmentation masks, amongst other tasks. The work's main contribution is the introduction of shortcut or skip connections, which we will discuss in more detail below.

\begin{figure}
 \centering
 \includegraphics[width=1\textwidth]{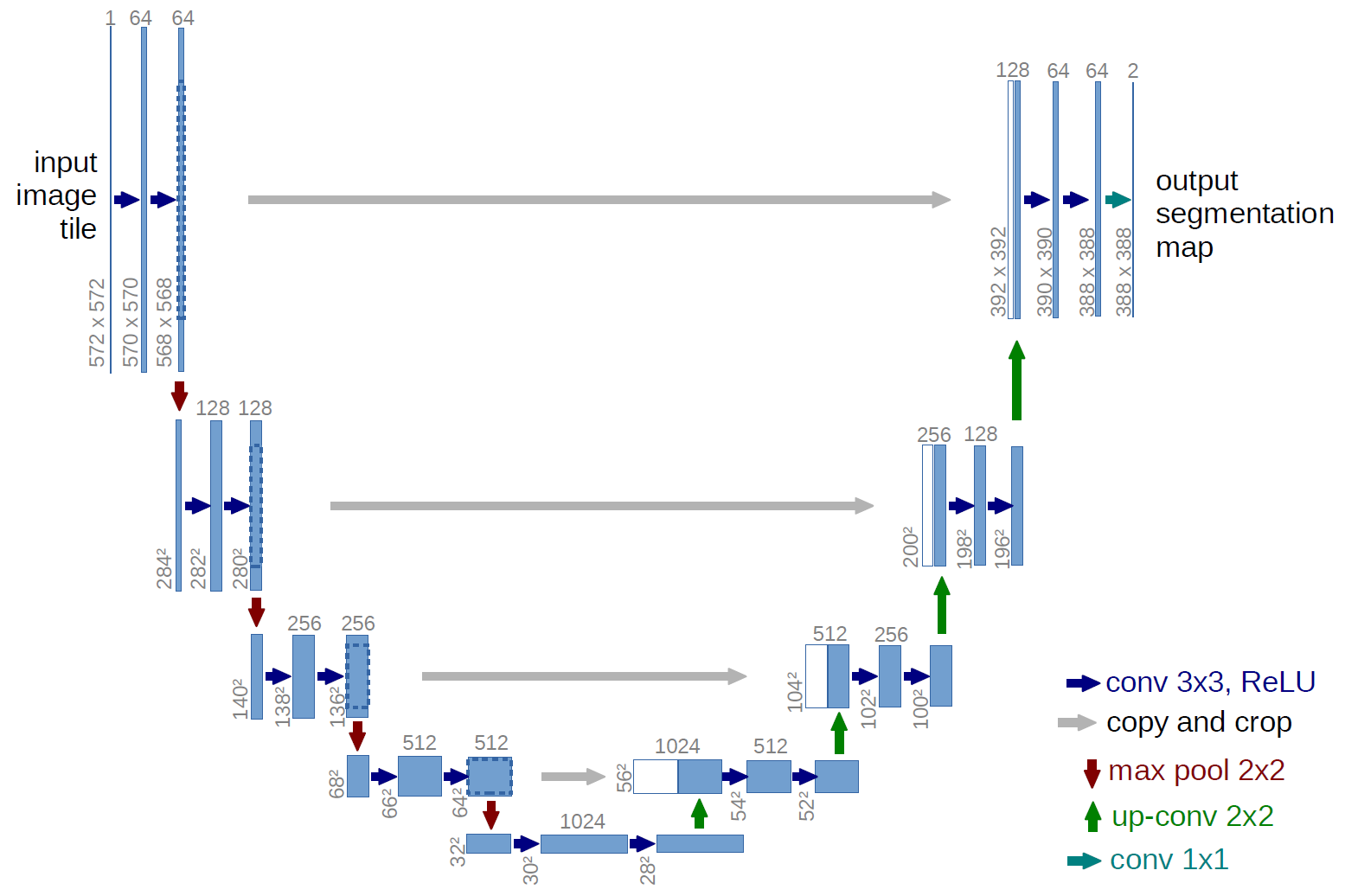}
 \caption{U-net architecture (example for 32x32 pixels in the lowest resolution). Each blue box corresponds to a multi-channel feature map. The number of channels is denoted on top of the box. The x-y-size is provided at the lower left edge of the box. White boxes represent copied feature maps. The arrows denote the different operations. Adopted from \citep{Ronneberger2015U-net:Segmentation}}
 \label{fig:u-net_architecture}
\end{figure}

The first half of the u-net is just a traditional stack of convolutional and max pooling layers (the encoder in Figure \ref{fig:u-net_architecture}). This half takes care of finding ``what" is in the image, just like a regular CNN finds characteristics in the image from edges (in earlier layers) to more complex structures (in deeper layers). The second path is the symmetric expanding path (the decoder in Figure \ref{fig:u-net_architecture}) which is used to enable precise localisation using transposed convolutions. This prevents the ``where" information of objects from getting lost by gradually applying up-sampling. Moreover, at every step of the decoder, u-nets use skip connections denoted by the grey arrows in Figure \ref{fig:u-net_architecture}. These skip connections concatenate the output of the transposed convolution layers with the feature maps from the Encoder at the same level. Skip connections are very powerful in the decoding process. A decoder with skip connections allows the network to adjust coarse predictions using lower-level features such as edges and other small patterns. Thus it is an end-to-end fully convolutional network (FCN), for example it only contains convolutional layers and does not contain any dense layers. All further details on u-nets can be found in \cite{Ronneberger2015U-net:Segmentation}.

\subsection{Challenges when segmenting aerial images}
\begin{figure}[H]
 \centering
 \includegraphics[width=1\textwidth]{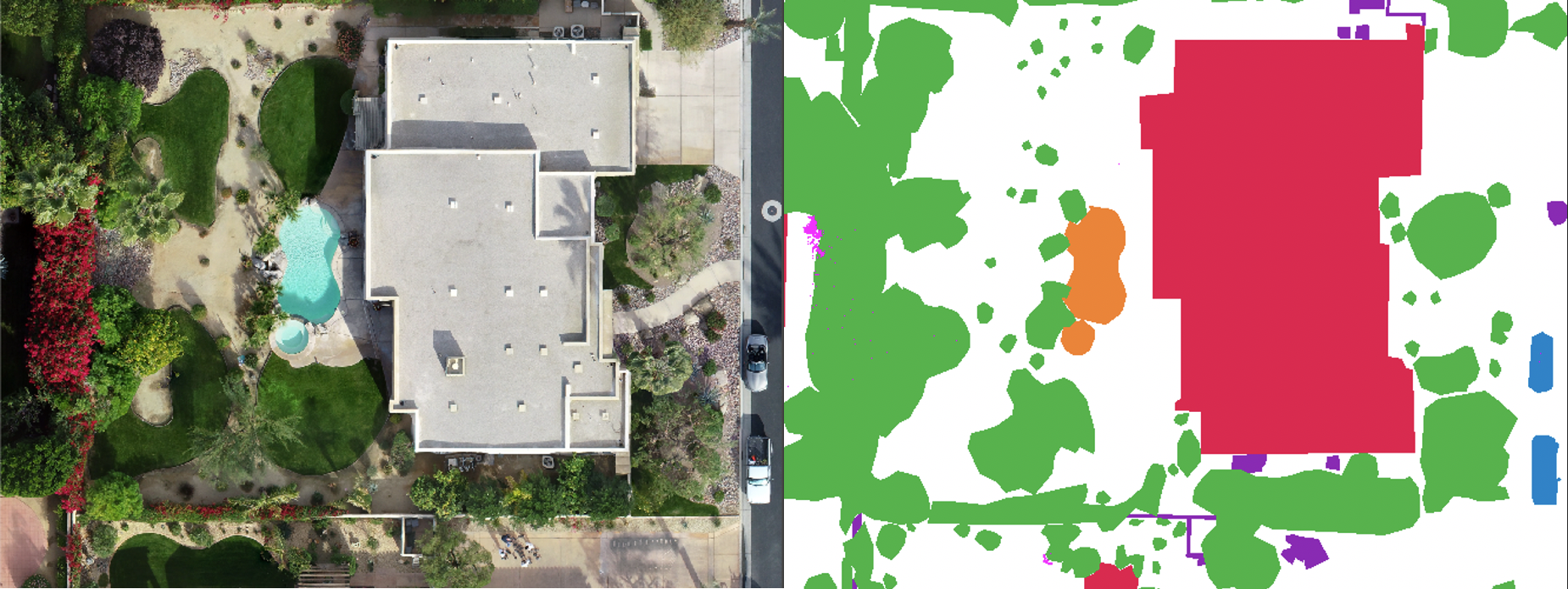}
 \caption{A partial image of the DroneDeploy dataset \citep{Pilkington2019GitHubBenchmark} on the left with 5 different classes, together with its corresponding segmentation mask on the right.}
 \label{fig:dronedeploy-classes}
\end{figure}

Segmenting datasets with one (\cite{Shermeyer2020SpaceNetDataset}, \cite{Soman2019RooftopImagery}) or a few classes \citep{Sang2018FullySegmentation} has been done using u-nets or similar performing architectures, but aerial imagery's biggest potential is the ability to capture a large portion of the earth (with a larger number of classes) at the same time. To illustrate this, we present a partial image taken from the DroneDeploy dataset used for this research in Figure \ref{fig:dronedeploy-classes}. 

However, when compared to popular multi-class segmentation datasets such as CityScapes, Camvid and PASCAL VOC, there are two challenges. Namely, smaller image resolution per class and, since all pictures are taken from above in aerial imagery, there is only one image perspective for all classes. Furthermore, there are hardly any high quality labeled aerial imagery datasets with many classes available. Right now, physically smaller classes (for example cars, persons) cause accuracy issues and class imbalance, presumably because small classes give the network a relatively small number of pixels to learn from and to do inference on.

The challenge of aerial images' resolution gets amplified by the fact that the classes we are trying to predict in an image contain less pixels than the classes in traditional ground-level images. This is where FCN's and u-nets segmentation masks lack quality, because of excessive downsizing due to consecutive pooling operations. The performance of several FCN-based architectures is often still mentioned for reference in other works, such as the work from \cite{Chen2018DeepLab:CRFs}.

\subsection{Novel techniques to achieve pixel-level segmentation on aerial images}\label{sec:noveltechniques}
The next sections will elaborate on different more novel techniques which focus on accuracy. The model that we elaborate on the most is DeepLabv3+ \citep{Chen2018Encoder-decoderSegmentation}, because of their state-of-the-art performance on segmentation datasets, as well as the fact that the proposed techniques also seem suitable for aerial imagery challenges. We will explain these techniques in the coming sections. 

Other novel solutions were considered but not deemed ideal, such as YOLOv4 \citep{Bochkovskiy2020YOLOv4:Detection} which focuses on high-speed object detection and HRNet-OCR \citep{Tao2020HierarchicalSegmentation} which focuses quite specifically on the CityScapes dataset and its shortcomings regarding coarse labeling. Another very interesting work rethinks self-training, a semi-supervised learning concept which \cite{Zoph2020RethinkingSelf-training} use to show state-of-the-art segmentation results on the PASCAL VOC dataset. Since this work only came out in June 2020, we were not able to incorporate it into this work in time, but encourage the reader to experiment with this technique in future work.

\subsubsection{Dilated Convolution}\label{sec:dilated-conv}
Pooling operations are useful for extracting sharp low-level features such as edges and points. It is also done to reduce variance by letting information spread from any point in the image, and to reduce computational costs. However, when several pooling operations are used downwards in the encoder, a lot of fine-grained information is lost. Moreover, being a fully convolutional network for a segmentation problem, the decoder needs to upsample the features by the same factor using deconvolution which again is a memory and computational expensive operation because of the large number of learnable parameters.

The original DeepLab paper \citep{Chen2018DeepLab:CRFs} proposes the usage of dilated convolution which helps processing a larger context image using the same number of parameters. In theory this should be ideal for aerial images and their large coverage, although there is a trade-off between a large field-of-view and accurate localisation. We address this trade-off with Dilated Spatial Pyramidal Pooling, explained in Section \ref{sec:dspp}.

\begin{figure}
 \centering
 \includegraphics[width=1\textwidth]{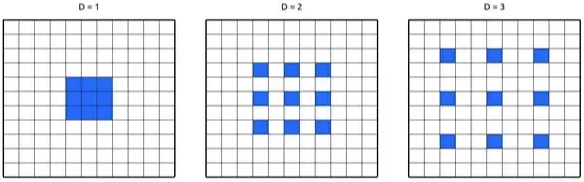}
 \caption{Dilated convolution filters with dilation rates $D=1$, $D=2$, $D=3$ respectively which correspond to the number of zeroes between the original pixel inputs. Original image from \citep{Artacho2019WaterfallSegmentation}}
 \label{fig:dilated}
\end{figure}

Dilated convolution increases the size of the convolution filter by appending zeros to fill the gaps between pixel inputs.  When the dilation rate (denoted in Figure \ref{fig:dilated} as $D$) $D=1$, it is equivalent to a conventional convolution. When $D=2$, a zero is inserted between every pixel input, making the filter cover a larger area. It now has the capacity to grasp the context of a 5x5 convolution filter, while having 3x3 convolution filter computational complexity. The center image in Figure \ref{fig:dilated} illustrates this. During up sampling a similar advantage regarding lower computational complexity holds, since bi-linear up sampling does not need any parameters as opposed to deconvolution. However, if we want to capture feature responses at multiple scales, we need to do multiple passes on one image. This negatively affects performance. The next technique will mitigate this issue. 

\subsubsection{Dilated Spatial Pyramidal Pooling}\label{sec:dspp}
The second technique proposed in \citep{Chen2018DeepLab:CRFs} is called Atrous Spatial Pyramidal Pooling (ASPP), which we will call Dilated SPP to remain consistent. The underlying technique (SPP) is first introduced in SPP-net by \cite{He2014SpatialRecognition}. One main contribution of SPP is very applicable to aerial imagery, namely robustly segmenting images from different scales and sizes. The spatial pyramidal pooling layers are able to capture information at multiple scales, by concatenating them to a 1-dimensional vector. In short, it works as follows. The Dilated CNN score map from each filter with different dilation rates is extracted and fused, by taking at each position the maximum response across the different scales. This significantly improves performance at minimal cost, because the feature responses of all Dilated CNN layers for multiple scales are computed from one original image, instead of processing the original image multiple times at different scales. In theory this contribution greatly benefits our goal, because aerial imagery is collected from a vast number of different heights and resolutions. 

\begin{figure}
 \centering
 \includegraphics[width=1\textwidth]{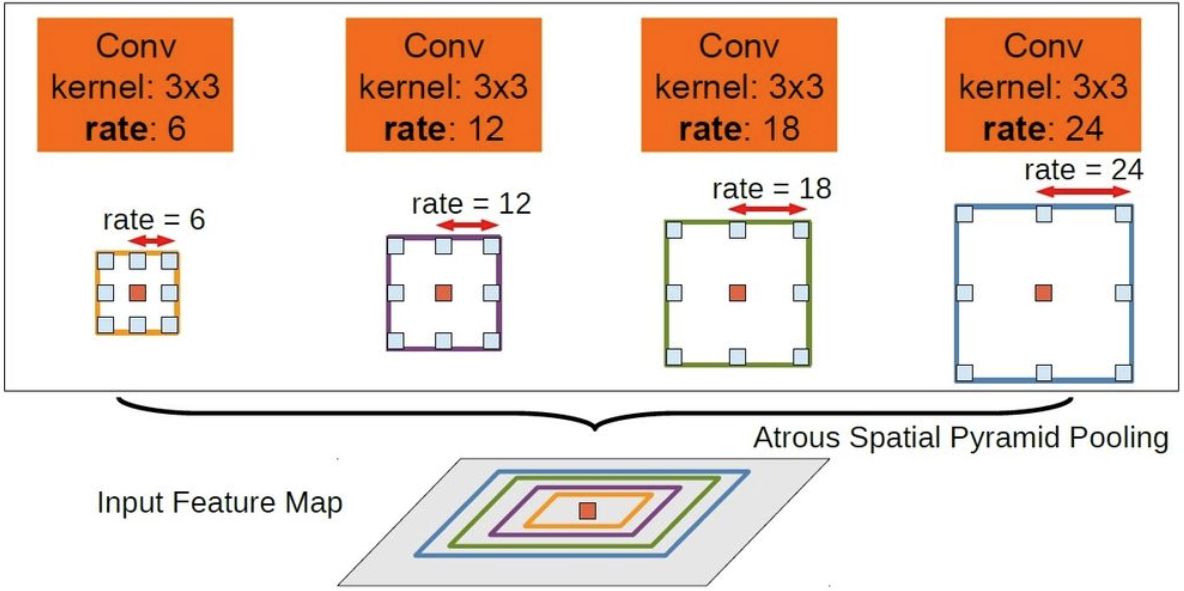}
 \caption{Dilated Spatial Pyramid Pooling with filter size 3 and dilation rates $D=6$, $D=12$, $D=18$, $D=24$ to create multiple Field-Of-Views shown in different colors. The extracted features from different dilation rates are then concatenated by converting to a 1d vector to create the final result. Original image from \citep{Chen2018DeepLab:CRFs}}
 \label{fig:aspp}
\end{figure}

Dilated SPP elegantly combines this technique with the same concept as dilated convolutions. A dilated convolution filter runs over the image using multiple dilation rates, after which SPP combines these filters to create one large feature map to ultimately classify the pixel of interest. Figure \ref{fig:aspp} shows this process in a 2-dimensional plane. 
To conclude, DSPP allows us to grasp a much larger context while classifying each pixel, whilst keeping the computational cost relatively low due to the use of dilated convolutions. Combining the techniques discussed in sections \ref{sec:fcn-unet}, \ref{sec:dilated-conv} and \ref{sec:dspp} we now have a computationally efficient convolution technique which is able to grasp a large context whilst classifying each pixel. However, due to the use of pooling layers and down-sampling, we lose locality accuracy. This is a problem, because the segmentation output will be coarse and the boundaries are not concretely defined. Especially with aerial images, where every pixel matters, this makes a difference. 

\subsubsection{Conditional Random Fields}\label{sec:crfs}
The third and final contribution of \citep{Chen2018DeepLab:CRFs} is the introduction of a fully connected Conditional Random Field (CRF). In more traditional applications, short-range CRF's are used to make noisy segmentation maps smoother, ``favoring same-label assignments to spatially proximal pixels". Instead, the authors use a fully connected CRF model which uses an energy function shown below in equation \ref{eq:energy}.
\begin{equation}\label{eq:energy}
    E(\mathbf{x}) = \sum_{i} \theta_i(x_i) + \sum_{ij} \theta_{ij}(x_i, x_j)
\end{equation}
The vector $\mathbf{x}$ is the label assignment for pixels. The first sum contains the unary potential $\theta_i(x_i) = -log P(x_i)$, where $P(x_i)$ is the label assignment probability at pixel $i$ computed by a Dilated CNN. The second sum contains the pairwise potential  $\theta_{ij}(x_i, x_j)$, which uses two Gaussian kernels in different feature spaces. The first kernel depends on both pixel positions and RGB color, the second kernel only depends on pixel positions. Together, these kernels similarly label pixels with similar color and position. For more details on the pairwise potential we refer the reader to the original Deeplab paper by \cite{Chen2018DeepLab:CRFs}. The essential benefit of using the fully connected CRF is that accurate (sharper) inference can now be done more efficient, especially on the boundaries of an object. 

Still, better results are possible by using the final adaptation of DeepLabv3+, the modified aligned xception backbone instead of a ResNet backbone.

\subsubsection{Modified Aligned Xception}\label{sec:xception}
As discussed in \citep{Chen2018Encoder-decoderSegmentation}, the authors of \citep{Dai2017DeformableNetworks} have shown promising results in the task of object detection using their Modified Aligned Xception model. Their main two contributions are deformable convolution and deformable Region of Interest (RoI) pooling, pushing the state-of-art results using the techniques discussed in sections \ref{sec:dilated-conv}, \ref{sec:dspp} and \ref{sec:crfs} even further. The main idea of deformable convolution is to make it possible to alter the common square shape of filters. To achieve this, the regular sample matrix in each filter/kernel is augmented with learnable offset data. This makes it possible to create specifically shaped filters, depending on which features the network is learning while predicting each class. 

\begin{figure}[H]
 \centering
 \includegraphics[width=1\textwidth]{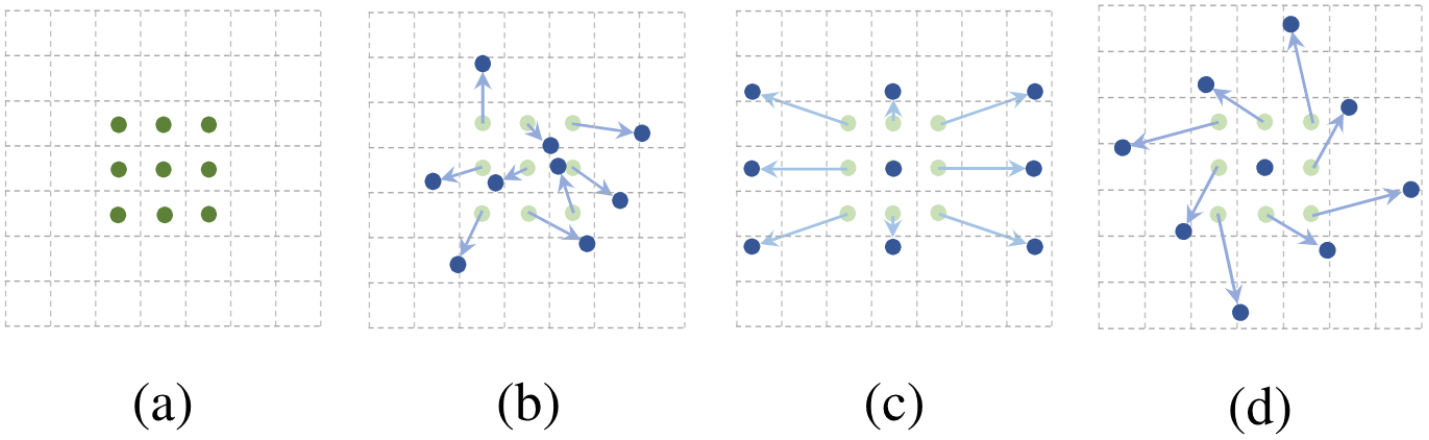}
 \caption{Illustration of the sampling locations in 3x3 standard and deformable convolutions. (a) regular sampling grid (green points) of standard convolution. (b) deformed sampling locations (dark blue points) with augmented offsets (light blue arrows) in deformable convolution. (c)(d) are special cases of (b), showing that the deformable convolution generalizes various transformations for scale, (anisotropic) aspect ratio and rotation. Adopted from \citep{Dai2017DeformableNetworks}}
 \label{fig:deform-conv}
\end{figure}

As visible in Figure \ref{fig:deform-conv}, deformable convolutions allow for more flexible sampling locations while using standard 3x3 sized filter parameters, albeit augmented with offset values. For the scope of this research, we choose to follow the same modifications to Xception as DeepLabv3+ in \citep{Chen2018Encoder-decoderSegmentation} since that configuration gave them the state-of-the-art results on PASCAL VOC 2012.
Having discussed the four techniques above, we will now compare these modern techniques against the current level of aerial image segmentation solutions used in other works.

\subsection{State-of-the-art of aerial image segmentation}\label{sec:sota}
A recently published journal article \citep{Wurm2019SemanticNetworks} in the International Society for Photogrammetry and Remote Sensing (ISPRS) used FCN-VGG19, without mentioning any of the architectures discussed above, or why they were not considered. Nevertheless, they state that they observed high performance of semantic segmentation for mapped slums in the optical data they used. Specifically, a mean IOU of 87.43\% over 4 different classes.

Another paper \citep{VietSang2018FullySegmentation} which proposes a deep learning approach to aerial image segmentation uses the somewhat outdated FCN architecture. They use 6 different classes, but only report precision, recall and F1 scores instead of the mean IOU which is a more conservative metric, especially if one of the classes scores relatively low which is the case for this work: clutter scores 34.1\% accuracy, but the overall reported score is 91.0\%.

Other recent work \citep{Wang2019AccurateImage} uses more modern architectures, but focuses on satellite imagery with just one class, either roads or buildings. They report a mean IOU of 65.1\% on the DEEPGLOBE road extraction dataset.

There is one work \citep{Liu2019AnDeeplabv3+} which uses the DeepLabv3+ architecture on the satellite imagery dataset SpaceNet-6 \citep{Shermeyer2020SpaceNetDataset}. However, this is also a single building-class dataset. Interestingly, in this work the authors state that the DeepLabv3+ architecture performs non-optimal on remote sensing images, due to the relatively simple decoder design. Therefore they add two extra upsampling layers and several skip connections between the encoder and the decoder. Using their decoder and the Xception encoder they outperform DeepLabv3+ by improving the mean IOU score from 73.1\% to 74.5\%. Since we discovered this work rather late during this research, we recommend additional experiments with this adapted DeepLabv3+ as a possibility for future work.

Considering the above mentioned state-of-the-art, we conclude that given a suitable dataset (for example with high enough quality), there is a lot of room for improvement on aerial imagery analysis in general, by using the current state-of-the-art architectures on richer datasets containing more than one class. This is exactly what we this work does, by applying the DeepLabv3+ architecture to the richer DroneDeploy dataset in Section \ref{sec:experiment-design}.

\section{Model design}\label{sec:model-design}
In this section we discuss the model design choices for our research for both u-net and DeepLabv3+. In Section \ref{sec:relatedwork} we covered each architectures' main contributions, advantages and disadvantages. We will now focus on their specific implementation in this section. Also we discuss several adaptations we had to make in order to create a compatible and complete implementation for our experiments. For a complete list of these alterations we refer the reader to Appendix B.

The u-net architecture discussed in Section \ref{sec:fcn-unet} is a well known and often used baseline in many dataset benchmark comparisons, as is the case for our research. More specifically, we use two different deep learning libraries with their own u-net implementation. One from Keras and the other from fastai's U-Net learner \citep{Howard2020Fastai:Learning} which is based on PyTorch. Fastai's u-net implementation uses PixelShuffle and sub-pixel convolution ``initialised to convolution NN resize" (ICNR) initialisation \citep{Aitken2017CheckerboardResize} on top of the regular u-net implementation and call this a ``DynamicUnet". PixelShuffle and ICNR combined allow for more modelling power at the same computational complexity and produces clean outputs by eliminating checkerboard artifacts. 
For the scope of this work it is not necessary to elaborate more on these techniques, since these are not used in any of our further experiments except for the fastai u-net implementation which we did not alter.

The main contributions of the DeepLabV3+ architecture has been thoroughly discussed in sections \ref{sec:dilated-conv}, \ref{sec:dspp} and \ref{sec:crfs}. The architecture shows state-of-the-art performance on all popular segmentation datasets mentioned in the introduction,  CityScapes (mIOU 82.1\% on test set), Camvid (mIOU 81.7\% \citep{Zhu2018ImprovingRelaxation}) and PASCAL VOC 2012 (mIOU 89\% on test set). Since one of the u-net benchmark implementations is based on PyTorch, we first attempted to use a PyTorch implementation for DeepLabv3+ as well.

Adapting the PyTorch version's code of DeepLabv3+ to our dataset turned out to be too comprehensive. The code base was built up quite modular in terms of running several experiments with different architectures, but not modular at all in terms of using it on new or custom datasets. For this reason, we turned to the official DeepLabv3+ implementation from Tensorflow which we were able to adapt to work well with our aerial dataset. The link to the original GitHub repository is available in \cite{Chen2018Encoder-decoderSegmentation}.

For all architectures (Keras u-net, PyTorch fastai u-net, Tensorflow DeepLabv3+) we added an additional focal loss function which is described in Section \ref{sec:loss}. Furthermore, since the original DeepLabv3+ implementation did not include our aerial DroneDeploy dataset, we made several changes to the data pre-processing code in order to make the DeepLabv3+ pipeline compatible for our work. For details on these changes and more information to recreate our research setup we refer the reader to Appendix A. 

\subsection{Generalisation techniques}
Improving the generalisation ability of deep learning models is a difficult challenge on its own. The better a model performs on unseen data, the higher its generalisation ability. Well performing strategies to improve this ability are for example (dropout) regularisation, batch normalisation, transfer learning using pre-trained networks and data augmentation. Regarding data augmentation, there is a slight difference in working with aerial imagery when compared to ground-level imagery. For instance, there is no added value in flipping an object over its y-axis if this variant does not occur in real life. Thus, y-axis flipping does not make sense for ground-level images but it does for aerial imagery. On the other hand, extensive zooming ($>5\%$) only seems makes sense for ground-level images, since from an aerial point-of-view the classes are seen from roughly the same distance, regardless of the dataset. In order to confirm this, we check this empirically by increasing zooming augmentation to $10\%$ and $15\%$ in Section \ref{sec:explore}.

\subsection{Encoders and backbones}
For both u-net implementations, our experiments on the dataset include many different ResNet encoders. The encoders we use are ResNet-18, ResNet-34, ResNet-50 and ResNet-101. The encoder weights are pre-trained on ImageNet and unfrozen. The decoder is not pre-trained in any way, as is usually the case. 

For the DeepLabv3+ implementation, our experiments on the dataset include two network backbones XCeption65 and ResNetV1-50 Beta. The modified ResNet-50 backbone replaces the first original 7x7 convolution with three 3x3 convolutions. The XCeption65 backbone uses the model initialisation \textit{xception65\_coco\_voc\_trainaug}, which is pre-trained on ImageNet, MS-COCO and PASCAL VOC 2012 train\_aug. The ResNetV1-50 Beta backbone uses the model initialisation \textit{resnet\_v1\_50\_beta\_imagenet} which is pre-trained on ImageNet and unfrozen. For both the u-net and DeepLabv3+ pre-trained ResNet encoders, using ImageNet is solely based on decreasing training time to obtain better results faster.

\subsection{Image tile input size}
Since aerial images are too large to fit into GPU memory as a whole, we slice the original images into smaller tiles. In order to gain insight in the effect of different tile sizes, we experiment with two different tile sizes on a subset of the aerial data. However, as stated we are bound by the limitations of the working memory of our GPU. Therefore we will perform the experiments with tile sizes 300x300 and 500x500 pixels respectively in Section \ref{sec:explore}.

\subsection{Evaluation metrics and loss functions}\label{sec:loss}
Since we are dealing with a pixel level segmentation problem, it might seem logical to think of pixel accuracy as a valid evaluation metric. However, this is not the case, since high pixel accuracy does not give any guarantees regarding segmentation quality.
This issue is caused by class imbalance. When the class distribution of an image is extremely imbalanced, for example one class covers 98\% of the original image, and the model classifies all pixels as that class, 98\% of pixels are classified accurately. For aerial datasets such as the DroneDeploy dataset this problem is more important since the ground (37.7\%) and vegetation (10.43\%) classes are dominantly present, so this cannot be ignored. Instead, the performance is measured in terms of pixel intersection-over-union averaged across the 6 classes (mIOU), which does take class imbalance into account because we measure the accuracy of each class in the image separately. mIOU is used by the current state-of-the-art models and makes our results easier to compare against current and future work. 

In terms of loss functions, a popular choice in multi-class segmentation challenges is the categorical cross entropy (CCE) loss $\mathcal{L}_{CCE}$, which is depicted in equation \ref{eq:cce} below. Here, $\hat{y}_{ic}$ is the $i$-th scalar value of class $c$ in the model output, $y_{ic}$ is the corresponding $i$-th target value for class $c$, and output size $D$ is the number of scalar values in the model output. Being negative (as usual), we ensure that the loss decreases when the modeled and true distributions get closer to each other. The total number of classes is shown with the letter $C$.
\begin{equation}\label{eq:cce}
    \mathcal{L}_{CCE} = -\sum_{c=1}^{C}\sum_{i=1}^{D} y_{ic} \cdot \log \hat{y}_{ic}
\end{equation}
CCE loss is still widely used in the current state-of-the-art, but given the expected class imbalance issues in our dataset, we also inspect a loss function which deals explicitly with class imbalance. The need for this is also discussed in \citep{Long2014FullySegmentation} where they discuss weighting their loss ``for each output channel in order to counteract a class imbalance present in the dataset". 

Thus, we also run several experiments with Focal loss, introduced in \citep{Lin2017FocalDetection}. Focal loss is designed to address accuracy issues in situations where there is a large ratio (for example 1:1000) or more between the imbalanced classes. In this publication the authors show how they arrive at the proposed Focal loss starting from a binary cross entropy loss. We refer the interested reader to their work and do not repeat these notational steps. The variant the authors used in practice is shown with equation \ref{eq:focal} where $\alpha = 0.25$ on a scale of $0.1 \leq \alpha \leq 0.999$ and $\gamma = 2$ on a scale of $0 \leq \gamma \leq 5.0$ worked best in their experiments. 
\begin{equation}\label{eq:focal}
    FL(p_t) = -\alpha_t(1-p_t)^{\gamma} \log(p_t)
\end{equation}
Focal Loss relies on a large ratio between imbalanced classes, because the contribution each class makes to the loss depends on the number of correct classifications. The more correct classifications, the less important this class becomes in terms of the loss. Thus, the loss depends more on problematic classes. With $\gamma = 2$, an example classified with $p_t= 0.9$ would have a 100 times lower loss than regular cross entropy. This means that this example has a 100 times lower impact on our optimisation process. In comparison, an example classified with $p_t = 0.5$ only has a 4 times lower loss than regular cross entropy and its contribution is therefore relatively larger.

In this section we have described several model design choices and the adaptations we made in order to prepare them for our experiments. In the next section we will find out how the models perform in practice.
\newpage
\section{Experiment design}\label{sec:experiment-design}
Both architectures' implementations have been introduced in Section \ref{sec:model-design}. In this section we look into how these design choices affect their performance in detail by experimenting on an aerial imagery dataset. 

\subsection{Candidate datasets}
Geospatial datasets comes in a large number of different file formats, sizes, and schemes. They often require specific domain knowledge which complicate their use in machine learning. There are a few datasets with high quality annotations, although they are still scarce in 2020. We consciously chose the DroneDeploy dataset, mainly because of its high quality segmentation masks. We will continue by describing the characteristics of this dataset in Section \ref{sec:dronedeploy}. Three other datasets were considered and inspected by implementing a few exploratory experiments, namely SpaceNet-6 \citep{Shermeyer2020SpaceNetDataset}, Defence Science \& Technology Laboratory Satellite Imagery Feature Detection (DSTL Kaggle challenge) and SkyScapes-Dense \citep{MajidAzimi2019SkyScapes-Fine-GrainedScenes}. However, these were not selected eventually due to the lack of multiple classes, labeling quality and public availability, respectively.

\subsection{The DroneDeploy dataset}\label{sec:dronedeploy}
The full DroneDeploy dataset \citep{Pilkington2019GitHubBenchmark} contains 55 RGB images, 9.1 GB in total. The individual image sizes vary from 6.8 MB to 637 MB, all with $0.1$ meter per pixel spatial resolution, taken from all over the United States along with elevation maps and segmentation masks. Since these images are too large to process at once, each one is sliced into tiles. The dataset is split into 35/8/12 images for train/validation/test purposes respectively. For purposes of being reproducible we use the same splits as documented in the dataset benchmark available at \citep{Pilkington2019GitHubBenchmark}. Depending on the tile resolution, the train and validation files combined (43 in total) contain either 1968 tiles of 500x500 pixels, or 6888 tiles of 300x300 pixels, all non-overlapping. The test images are only sliced during inference. 

Once the tiles are created and split into train and validation datasets, we are free to randomise the tile order. Furthermore, since this is a pixel-level segmentation task, there are no typical "boundary issues", which do occur in other tasks such as object detection if an object is only half present in a sliced tile.

The segmentation masks are annotated with 6+1 classes - namely Building, Clutter, Vegetation, Water, Ground, Car and ‘Ignore’ - where the 'Ignore' class refers to mask areas of missing labels or image boundaries. During the slicing process into tiles, all individual tiles containing only 'Ignore' pixels are left out of the rest of training and inference process. During this research we only use the 55 RGB images and their segmentation masks, since we attain to achieve a result which generalizes well, and will therefore hold for many different datasets. Other datasets, such as SpaceNet-6 and DSTL, also have image data in bands specific to satellite data, for example panchromatic, multispectral or SWIR. The distribution of all classes and their corresponding color map is provided below in Table \ref{tab:classes}. 

\begin{figure}[H]
 \centering
 \includegraphics[width=1\textwidth]{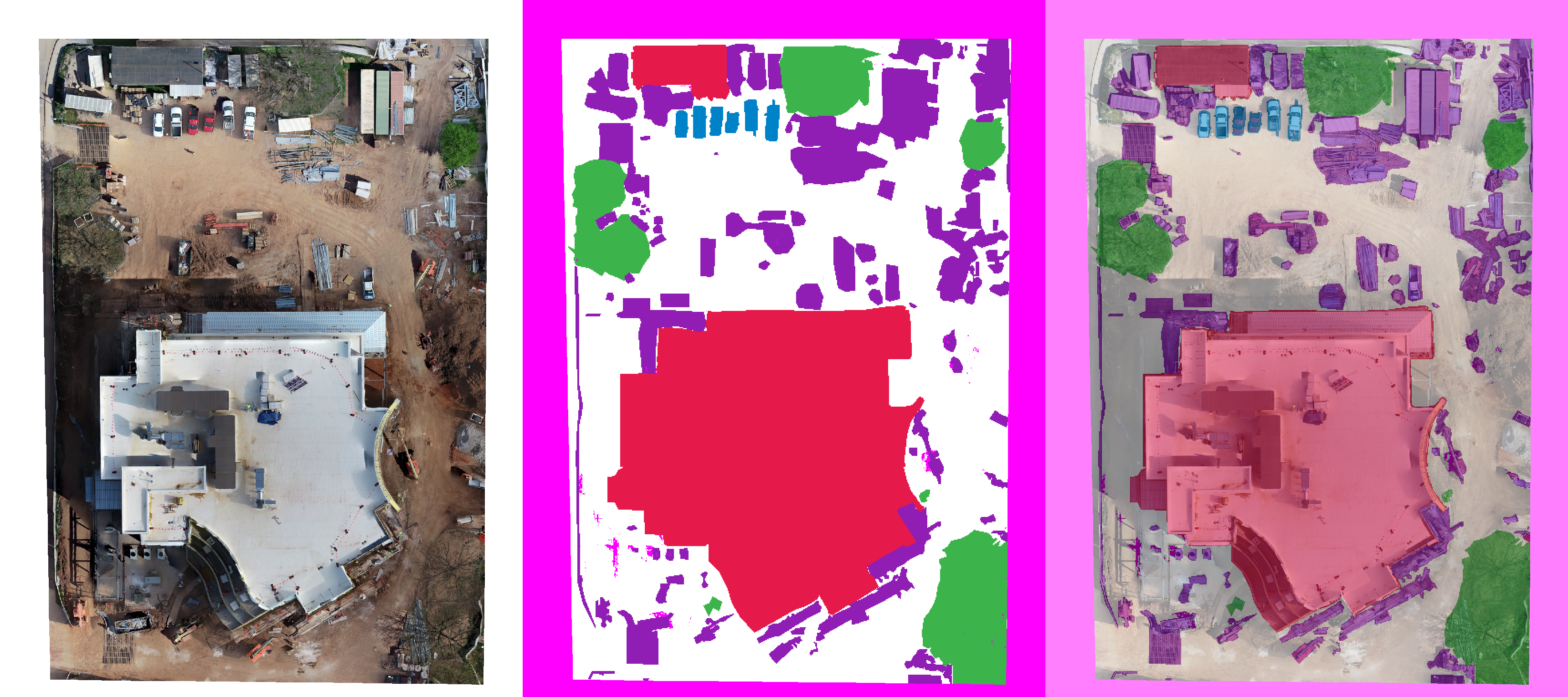}
 \caption{Image \textit{551063e3c5\_8FCB044F58INSPIRE} of the DroneDeploy dataset on the left, with its corresponding ground truth segmentation mask in the center and overlaid with 50\% transparency on the right}
 \label{fig:dronedeploy-example}
\end{figure}

\begin{table}[H]
\centering
\begin{tabular}{|l|l|l|}
\hline
\textbf{Class} & \textbf{Color code} & \textbf{Percentage of total pixels} \\ \hline
1: Building       & Red \textcolor{red}{\rule{0.2cm}{0.2cm}}                 & 5.6\%                               \\ \hline
2: Clutter/debris        & Purple \textcolor{purple}{\rule{0.2cm}{0.2cm}}             & 2\%                                 \\ \hline
3: Vegetation     & Green \textcolor{green}{\rule{0.2cm}{0.2cm}}              & 10.43\%                             \\ \hline
4: Water          & Orange \textcolor{orange}{\rule{0.2cm}{0.2cm}}             & 1.2\%                               \\ \hline
5: Ground         & White \textcolor{white}{\rule{0.2cm}{0.2cm}}           & 37.7\%                              \\ \hline
6: Car            & Blue \textcolor{blue}{\rule{0.2cm}{0.2cm}}               & 0.38\%                              \\ \hline
0: Ignore         & Magenta \textcolor{pink}{\rule{0.2cm}{0.2cm}}            & 42.7\%                              \\ \hline
\end{tabular}
\caption{Pixel distribution of the 6 annotated classes + 1 Ignore class in the DroneDeploy dataset}
\label{tab:classes}
\end{table}
Figure \ref{fig:dronedeploy-example} shows one full image and its corresponding ground truth segmentation mask. We can see cars, ground, buildings, vegetation, clutter and a large number of ignore pixels. Most of the images are largely labeled like this one. Again, sliced tiles containing ignore pixels only are intentionally left out of the training process.

\subsection{Performance exploration}\label{sec:explore}
While exploring design choices, we explore different choices in terms of encoders, augmentation, normalisation and loss functions before we go ahead and train a configurated network on the full dataset. The DroneDeploy dataset GitHub repository implementation uses the u-net architecture, based on either Keras or fastai (PyTorch). First we establish a baseline with this implementation and obtain mean F1-scores of $0.7806$ and $0.8026$ using fastai and Keras u-nets respectively, which is trained for 15 epochs. These results are also displayed in Table \ref{tab:results}.

\begin{figure}
 \centering
 \includegraphics[width=1\textwidth]{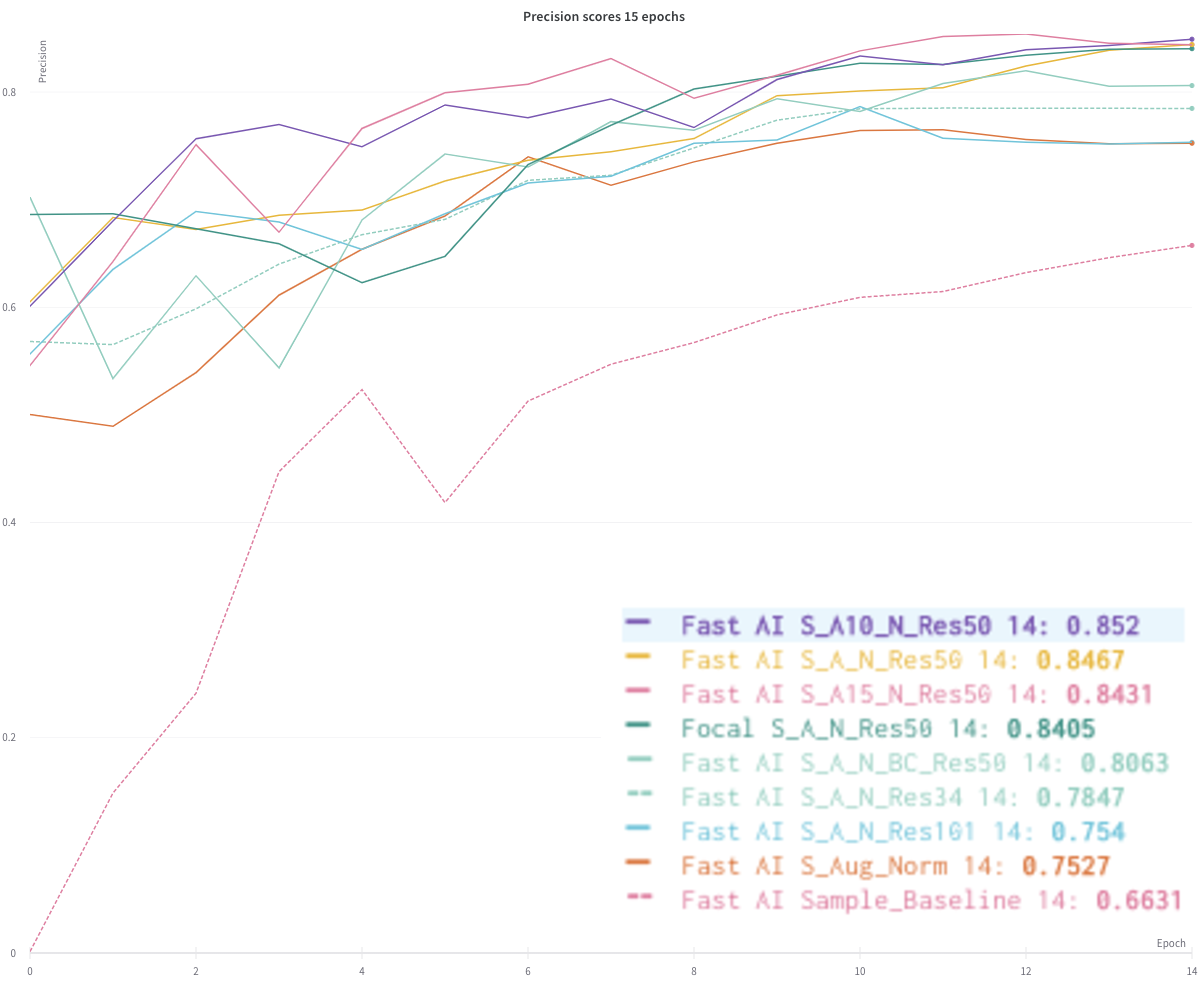}
 \caption{Precision scores recorded using the Weights and Biases package, for all exploratory alterations while training the DroneDeploy sample dataset. \textbf{S} : Sample, \textbf{A} : Augmentation with 5\% zoom, \textbf{A10/15} : Augmentation with 10\% or 15\% zoom respectively, \textbf{N} : Batch Normalisation, \textbf{BC} : Bigger tile size (500x500). \textbf{ResXX} denotes the used encoder and \textbf{Focal} denotes using the Focal Loss together with the Fast AI u-net implementation. Added final scores after 15 epochs in bottom right for readability in non-digital form.}
 \label{fig:precision}
\end{figure}

\begin{figure}
 \centering
 \includegraphics[width=1\textwidth]{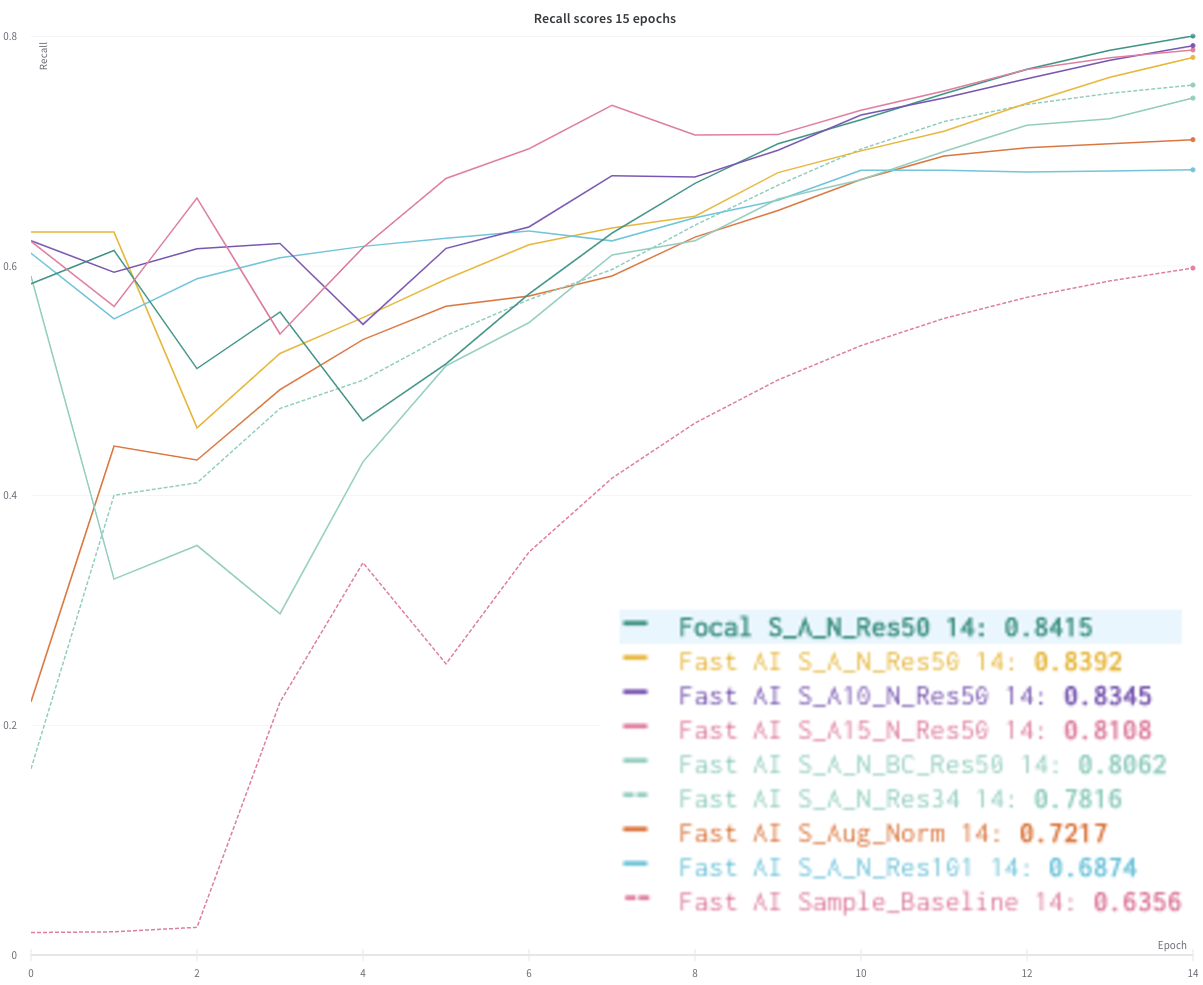}
 \caption{Recall scores recorded using the Weights and Biases package, for all exploratory alterations while training the DroneDeploy sample dataset. \textbf{S} : Sample, \textbf{A} : Augmentation with 5\% zoom, \textbf{A10/15} : Augmentation with 10\% or 15\% zoom respectively, \textbf{N} : Batch Normalisation, \textbf{BC} : Bigger tile size (500x500). \textbf{ResXX} denotes the used encoder and \textbf{Focal} denotes using the Focal Loss together with the Fast AI u-net implementation. Added final scores after 15 epochs in bottom right for readability in non-digital form.}
 \label{fig:recall}
\end{figure}

Next, we explore encoders, tile sizes, augmentation, normalisation and loss function options, as discussed in Section \ref{sec:model-design}, by using the \textit{dataset-sample} DroneDeploy subset. In an iterative process, we measure the precision and recall of each implementation after 15 epochs on the sample dataset and base our decision on the interim results. Precision and recall scores are displayed in Figures \ref{fig:precision} and \ref{fig:recall} respectively. Unless specified otherwise, the tile size is always 300x300 pixels. The sample baseline is ran without any alterations, using a pre-trained ResNet18 encoder and categorical cross entropy loss.

While inspecting the precision and recall scores, we see that adding data augmentation, batch normalisation and deeper encoders generally helps performance. Experimenting with different zoom augmentation levels shows that increasing zoom from $5\%$ to $10\%$ does improve the precision with a marginal $0.0053$, while the recall drops with $0.0047$. The precision and recall scores both decrease when we increase zoom augmentation further to $15\%$. 

Using the alternate Focal loss function does not show immediate performance gain. This is not very surprising given the fact that the focal loss performs better when there is a large ratio class imbalance (see Section \ref{sec:loss}) which might not be the case for the sample dataset. To verify the performance of the Focal loss function we will include this variant in the full dataset experiments. 

Somewhat surprisingly, increasing the tile size to 500x500 pixels per tile has a negative effect on performance. This might be the result of overfitting, because the model is trained with more parameters on a smaller amount of tiles due to the tile size. While this might only be the case with the basic u-net architecture, we choose not to pursue this further, because of GPU memory limitations and because there does not seem to be a clear advantage. Lastly, in terms of encoders the ResNet50 architecture seems to hit the ``complexity sweet spot" for our task after testing with ResNet18, ResNet34, ResNet50 and ResNet101 encoders. We will use these preliminary results as pointers when training our architectures on the full dataset.

\subsection{Training the network}
For training purposes, the dataset is divided into non-overlapping tiles with tile sizes of 300x300 pixels in which at least 1 class of interest is present. 
The total dataset contains 6888 image-tiles. 14 image-tiles only contain the ``Ignore'' class, which we omit during training for this reason. Hence we have 6874 image-tiles at our disposal. Interestingly, there are 1524 image-tiles containing only one relevant class, which occurs for the classes ``Building'', ``Clutter'', ``Vegetation'' and ``Water''. Apart from ``Clutter'', which is a rather ambiguous name for a class, this seems to make sense.

Training is done using a single GPU, on either a GeForce GTX 1080 Ti with 11GB of RAM or a Tesla V100 with 16GB of RAM. The details of both clusters are shown in Table \ref{tab:hpc}.

\begin{table}[]
\begin{tabular}{|l|l|l|l|l|}
\hline
\textbf{CPU Type}                                                     & \textbf{RAM}    & \textbf{GPU Type}                       & \textbf{RAM}  & \textbf{CUDA cores} \\ \hline
Intel Xeon Broadwell-EP 2683v4 & 1024GB & GeForce GTX 1080Ti & 11GB & 3584       \\ \hline
\end{tabular}
\caption{Details of the cluster used for training and experimentation}
\label{tab:hpc}
\end{table}
\newpage
\subsection{Results}\label{sec:results}
\begin{figure}[H]
 \centering
 \includegraphics[width=1\textwidth]{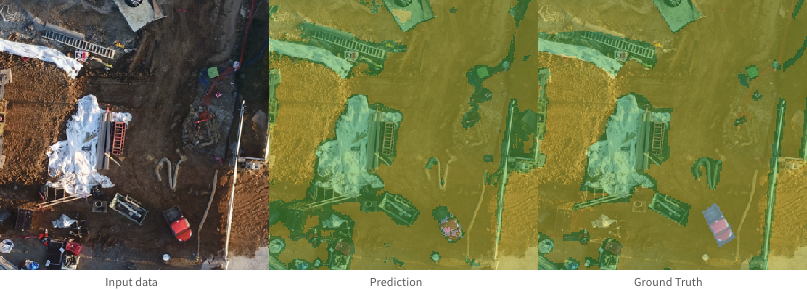}
 \caption{Zoomed in image of the DroneDeploy dataset on the left, fastai Focal ResNet50 u-net baseline's predicted semantic segmentation mask in the center and its ground truth on the right}
 \label{fig:u-netprediction}
\end{figure}

\begin{figure}[H]
 \centering
 \includegraphics[width=0.88\textwidth]{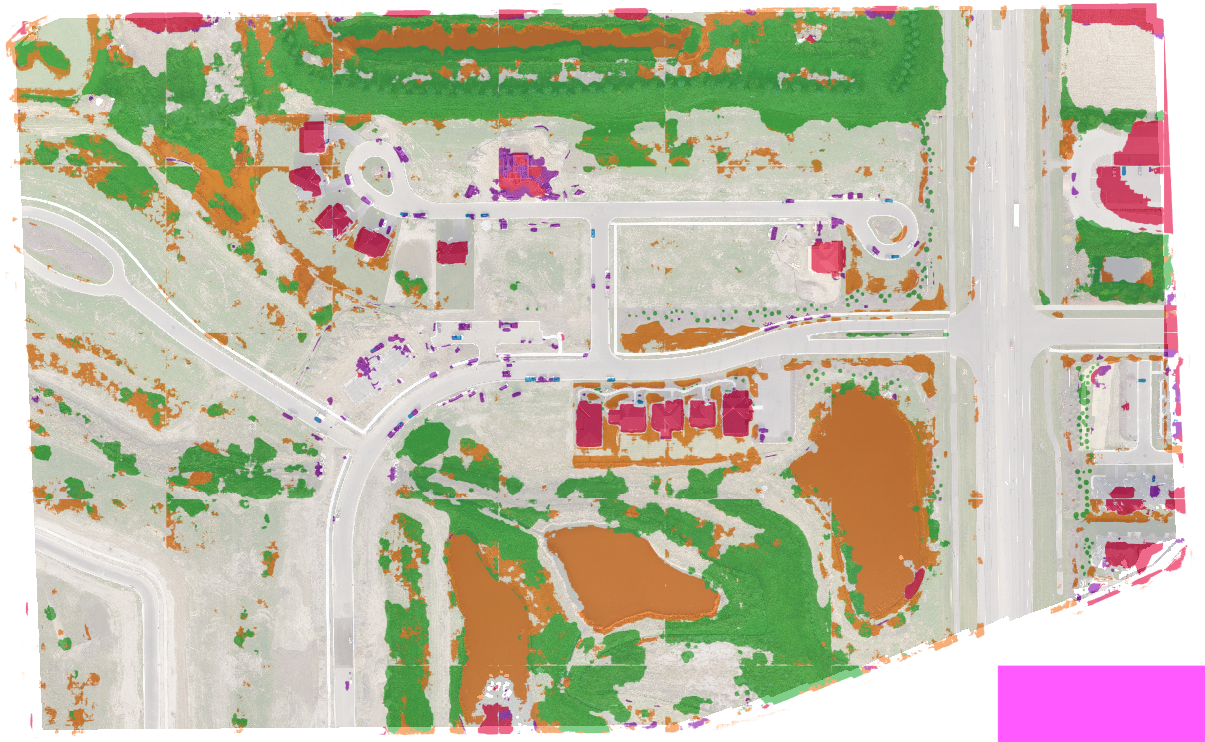}
 \caption{Image \textit{1476907971\_CHADGRISMOPENPIPELINE} of the DroneDeploy dataset overlaid with 65\% transparency predicted label by keras CCE ResNet50 u-net baseline}
 \label{fig:large-prediction}
\end{figure}

In this section we present our results on the DroneDeploy dataset in Table \ref{tab:results}. The DeepLabv3+ models are trained for 40 epochs on the train dataset and evaluated on the validation and test set. The u-net models are trained on the train dataset and tuned on the validation set. Hence, we only report the mIOU scores on the test set for these models. Figure \ref{fig:u-netprediction} gives an impression of the fastai Focal u-net model's performance, while Figure \ref{fig:large-prediction} shows the performance of the keras u-net CCE model performance. 

\begin{figure}[H]
 \centering
 \includegraphics[width=1\textwidth]{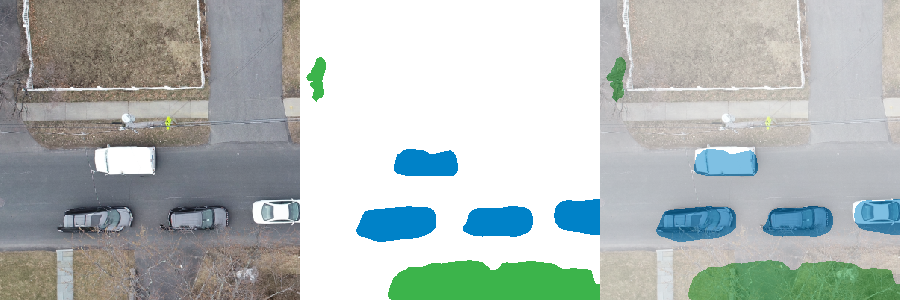}
 \caption{Showcasing the performance on the ``car'' class. Zoomed in image tile of the DroneDeploy test set on the left. In the center the DeepLabv3+ Xception65 prediction mask, with ground \textcolor{white}{\rule{0.2cm}{0.2cm}}, cars \textcolor{blue}{\rule{0.2cm}{0.2cm}} and vegetation \textcolor{green}{\rule{0.2cm}{0.2cm}} pixels. On the right, the prediction mask overlaid with 50\% transparency.}
 \label{fig:cars1}
\end{figure}

\begin{figure}[H]
 \centering
 \includegraphics[width=1\textwidth]{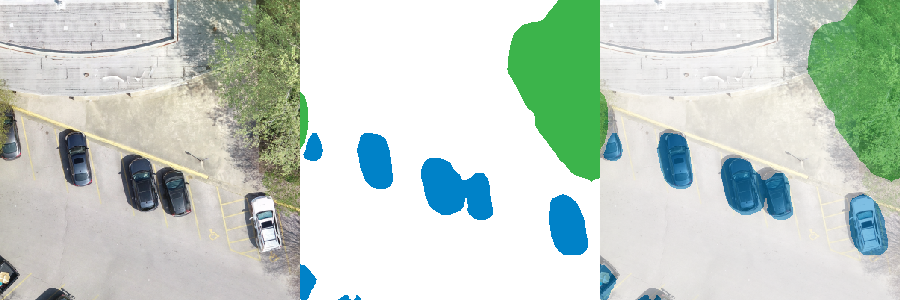}
 \caption{Showcasing the performance on the ``car'' class. A second zoomed in image tile of the DroneDeploy test set on the left. In the center the DeepLabv3+ Xception65 prediction mask, with ground \textcolor{white}{\rule{0.2cm}{0.2cm}}, cars \textcolor{blue}{\rule{0.2cm}{0.2cm}} and vegetation \textcolor{green}{\rule{0.2cm}{0.2cm}} pixels. On the right, the prediction mask overlaid with 50\% transparency.}
 \label{fig:cars2}
\end{figure}

In a very recent, unpublished preprint \citep{Parmar2020ExplorationDrones} the authors show a 65.0 mIOU score on the validation set. In this work there is no score reported on the test set. Our mIOU score using the DeepLabv3+ architecture of 69.9 is achieved on the same validation set, calculated over the first 5 classes. For the sixth class, ``car'', the DeepLabv3+ implementation consistently reports ``not a number". Nevertheless, as the prediction images in Figures \ref{fig:cars1} and \ref{fig:cars2} show, the model seems to perform quite well on cars as well. Nevertheless the prediction is not perfect yet, because the left most car in Figure \ref{fig:cars2} suffers from an overhanging tree which causes a--somewhat understandable--misclassification. Hence, with a small reservation for this class, we show the potential to outperform the current publicly available state-of-the-art mIOU with 4.9 percent. Furthermore, there is no other mIOU benchmark for the test set to our knowledge. Hence, we propose a new benchmark on the DroneDeploy test set using the best performing DeepLabv3+ Xception65 architecture, with a mIOU score of 52.5\%.


All current scores presented in Table \ref{tab:results} are obtained by training for 40 epochs on the GeForce GTX 1080Ti GPU with the aforementioned split of 35/8 for the training and validation data respectively. All final mIOU scores are measured on the 12 images from the test set.

\begin{table}[H]
\begin{tabularx}{\textwidth}{|l|X|X|X|X|X|X|l|l|l|}
\hline
\textbf{Method}                  & \textbf{1} & \textbf{2} & \textbf{3} & \textbf{4} & \textbf{5} & \textbf{6} & \textbf{f1} & \textbf{val} & \textbf{test} \\ \hline
DLv3+ (Xception65)          & 46.0              & 79.6             & 79.6                & 88.8           & 55.4            & nan          & -                 & \textbf{\underline{69.9}} & \textbf{\underline{52.5}}          \\ \hline
DLv3+ (ResNetV1\_50\_Beta)  & 48.6              & 67.6             & 85.7                & 88.4           & 47.3            & nan          & -                 & 67.5 & 48.0          \\ \hline
Keras u-net (ResNet50) Focal     & 36.9               & 17.7              & 56.1                 & 15.9            & 73.9             & 36.1          & 0.8053            & - & 39.8          \\ \hline
Keras u-net (ResNet50) CCE       & 34.2               & 17.6              & 53.5                 & 18.7            & 75.6             & 52.8          & 0.8144               & - & 42.5           \\ \hline
fastai u-net (ResNet50) Focal & 28.3               & 8.2              & 37.0                 & 7.2            & 71.7             & 28.8          & 0.7536            & - & 30.2            \\ \hline
fastai u-net (ResNet50) CCE & 30.2               & 8.3              & 43.8                 & 5.6            & 66.7             & 21.5         & 0.7387            & - & 29.4            \\ \hline
\end{tabularx}
\caption{Results of DeepLabv3+ and u-net architecture variants trained on the DroneDeploy dataset. \newline \textbf{val}: mean IOU on validation set trained on train set only. test: mean IOU on test set trained on train+validation set. \newline \textbf{Class IOU scores}: 1:Building, 2:Clutter, 3:Vegetation, 4:Water, 5:Ground, 6:Car \newline \textbf{f1}: f1\_mean, reported for benchmark architecture comparison against DroneDeploy public leaderboard.}
\label{tab:results}
\end{table}

Additionally, we present our results on the DroneDeploy test set in Table \ref{tab:results-trainval}. Here however, the models are trained on both the train and the validation set for 40 epochs. 

\begin{table}[H]
\begin{tabular}{|l|l|l|l|l|l|l|l|}
\hline
\textbf{Method}            & \textbf{1} & \textbf{2} & \textbf{3} & \textbf{4} & \textbf{5} & \textbf{6} & \textbf{test} \\ \hline
DLv3+ (Xception65)         & 8.7              & 55.8             & 21.5                & 85.0           & 12.0            & nan          & 36.6          \\ \hline
DLv3+ (ResNetV1\_50\_Beta) & 0.3              & 54.1             & 12.3                & 86.2          & 0            & nan          & 30.6          \\ \hline
\end{tabular}
\caption{Results of DeepLabv3+ architecture variants trained on the DroneDeploy train and validation images. \newline \textbf{test}: mean IOU on test set. \newline \textbf{Class IOU scores}: 1:Building, 2:Clutter, 3:Vegetation, 4:Water, 5:Ground, 6:Car.}
\label{tab:results-trainval}
\end{table}
\begin{figure}[H]
 \centering
 \includegraphics[width=0.88\textwidth]{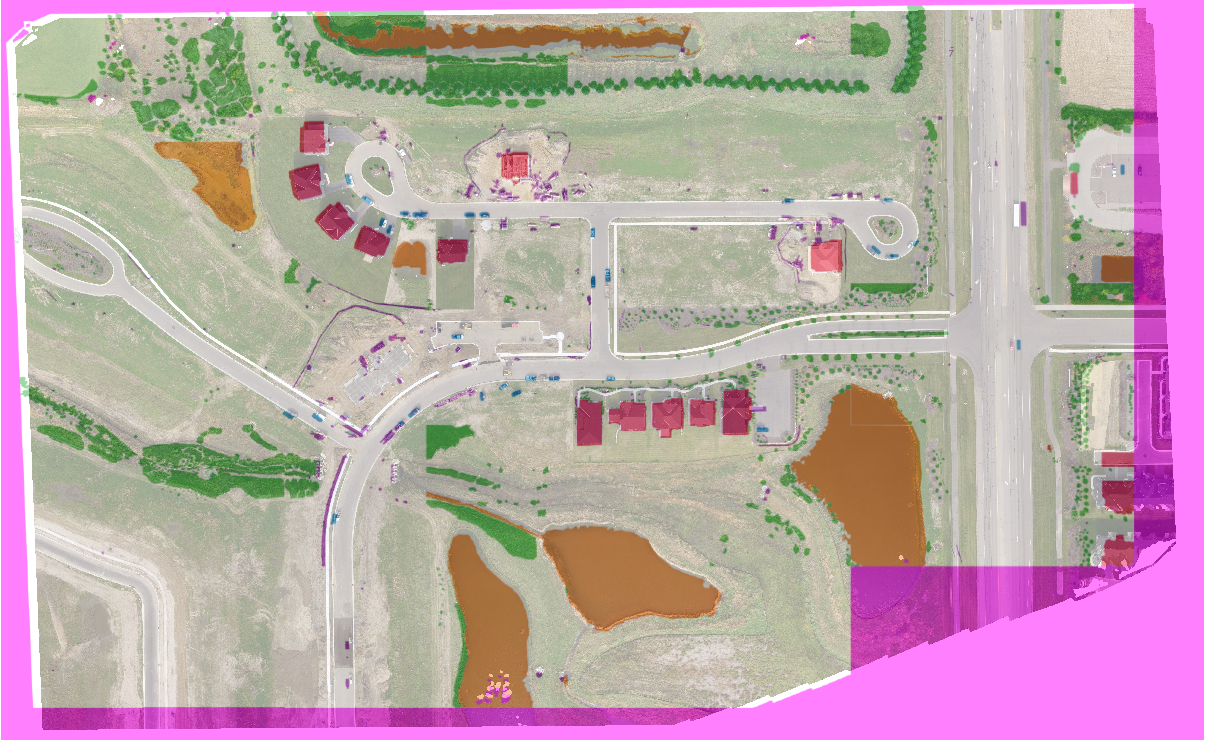}
 \caption{Image \textit{1476907971\_CHADGRISMOPENPIPELINE} of the DroneDeploy dataset overlaid with 50\% transparency ground truth label}
 \label{fig:large-groundtruth}
\end{figure}
\subsection{Discussion}
Surprisingly, the models trained on the train and validation set in perform worse on the test set than the models trained on just the train set. Since we used the given dataset split of 35/8/12 for respectively the train, validation and test set, we suspect that the original 55 RGB images were not thoroughly randomised. Specifically, we suspect that the validation set is significantly different than the test set, since the scores decrease when we add the validation set to our training data. Further investigation of these results is left as a recommendation for future work.

Also, while inspecting the model performance and comparing several images' ground truth mask with its corresponding prediction mask, we found that the ``ground truth'' labels can be debatable at times. In Figure \ref{fig:large-prediction} for example, there is a large patch of predicted ``vegetation'' between the orange labeled ``water'' pools at the bottom. In the ground truth mask in Figure \ref{fig:large-groundtruth} this is annotated as ``ground'', while it clearly shows characteristics of ``vegetation'' in the original image. The fact that roads and dirt also classify as ``ground'' strengthens the comment that a green patch, sharing more characteristics with ``vegetation'', should actually be classified as ``vegetation'' instead of ``ground''.

\section{Conclusions and future work}\label{sec:futurework}
We conclude this work by presenting our conclusions and recommendations for future work. The complete relevant codebase for this work is available online at \url{https://github.com/mrheffels/aerial-imagery-segmentation}. The other repositories we build upon are all mentioned in Appendix B.

\subsection{Conclusions}
In this paper, we present the performance of a fundamental segmentation neural network architecture, u-net, including a large number of different experiments while exploring their effect on the model performance. Based on the results of our exploration, we find that data augmentation, batch normalisation and the ResNet50 encoder generally help performance. Zoom augmentation above $5\%$, tile sizes of 500x500 pixels and the alternate Focal loss, all do not improve performance convincingly.

Finally, we propose a new performance benchmark for the DroneDeploy validation and test set. We obtain mean IOU scores of 69.9\% and 52.5\% on the validation and test set respectively, using the DeepLabv3+ Xception65 architecture.

\subsection{Future work}
For future work, we greatly encourage the interested researcher to extend the results found in this research for use on satellite imagery, for example to train networks more specifically using data in the panchromatic, multispectral or SWIR wavebands. Also, as stated in Section \ref{sec:sota}, a straightforward first extension can be made by altering the DeepLabv3+ architecture as proposed by \cite{Liu2019AnDeeplabv3+}.

Furthermore, it might be interesting to use this new baseline and extend it by incorporating the available elevation maps of the DroneDeploy dataset. In general, being able to use different types of input data rather than just RGB data can be of great value for aerial image machine learning problems. One can imagine that other input data such as elevations can help a network distinguish the difference between for example cars, ground or buildings a lot easier than when it is just using RGB input data.

As stated in Section \ref{sec:noveltechniques}, another potential extension is self-training, a semi-supervised learning concept which \cite{Zoph2020RethinkingSelf-training} use to show state-of-the-art segmentation results on the PASCAL VOC dataset. This very recent work also shows promising opportunities for aerial imagery, since the number of annotated examples are relatively low, which could be compensated using this technique in future works. Another argument for this extension's potential is the fact that high quality aerial imagery annotation is very time consuming to do manually. A well trained model has the potential to create new high quality annotations on a large scale.

Lastly, as stated in the discussion, if one does not take into account the given DroneDeploy dataset train/validation/test split, it is possible to reshuffle the complete dataset and experiment with a custom defined dataset split. For this work this was not within scope, to make the scores comparable to other public benchmarks.

\newpage
\appendix
\section*{Appendix A: creating environments for the DroneDeploy fastai/keras benchmark and DeepLabv3+ codebase}\label{app:envs}
It is crucial to take note of your own GPU driver environment. For this work the following environment was available (High Performance Cluster at Eindhoven University of Technology):
\begin{itemize}
    \item GCC \& G++: 5.4.0 (20160609)
    \item Nvidia driver: 450.51.06
    \item CUDA driver: 11.0
    \item CUDA compilation tools (including nvcc): release 10.2, V10.2.89
    \item Using TensorFlow through Conda automatically installs the latest CuDNN version in your local environment.
\end{itemize}
\subsection*{fastai environment and preparations}
\begin{enumerate}
    \item conda create --name fastai\_gpu pytorch torchvision cudatoolkit=10.1 -c pytorch
    \item conda activate fastai\_gpu
    \item conda install -c fastai fastai=1.0.61
    \item conda install opencv typing wandb scikit-learn
    \item pip install image-classifiers
\end{enumerate}

\subsection*{keras environment and preparations}
\begin{enumerate}
    \item conda create --name keras\_gpu keras tensorflow-gpu=1.15
    \item conda activate keras\_gpu
    \item conda install opencv typing wandb scikit-learn
    \item pip install image-classifiers
\end{enumerate}

\subsection*{DeepLabv3+ environment and preparations}
\begin{enumerate}
    \item conda create --name tf1\_gpu tensorflow-gpu=1.15
    \item conda activate tf1\_gpu
    \item conda install -c conda-forge pillow tqdm numpy
    \item pip install tf\_slim
    \item From tensorflow/models/research/ directory run: \\``export PYTHONPATH=\$PYTHONPATH:`pwd`:`pwd`/slim''
\end{enumerate}
\newpage
\section*{Appendix B: list of altered files for DroneDeploy dataset experiments}
The complete codebase for this work is also available online\footnote[1]{https://github.com/mrheffels/aerial-imagery-segmentation}.
\subsection*{DeepLabv3+ Tensorflow research codebase\footnote[2]{https://github.com/tensorflow/models/tree/master/research/deeplab} extensions}
\begin{itemize}
    \item convert\_rgb\_to\_index.py (altered to strip 3 dimensional segmentation labels to 1 dimensional)
    \item build\_dd\_data.py (altered for DroneDeploy compatiblity)
    \item data\_generator.py (altered for DroneDeploy compatiblity)
    \item train-dd-full.sh, eval-dd.sh, vis-dd.sh (dataset adaptations inspired by this GitHub repo\footnote[3]{https://github.com/heaversm/deeplab-training})
\end{itemize}
\subsection*{DroneDeploy benchmark codebase\footnote[4]{https://github.com/dronedeploy/dd-ml-segmentation-benchmark} extensions}
\begin{itemize}
    \item custom\_training.py (implementation Focal loss function for fastai u-net)
    \item custom\_training\_keras.py (implementation Focal loss function for Keras u-net)
    \item images2chips.py (added test images conversion to tiles for DeepLabv3+ compatibility)
    \item scoring.py (added mean IOU and IOU score per class metric)
\end{itemize}

\vskip 0.2in
\bibliography{main.bib}

\begin{thebibliography}{34}
\providecommand{\natexlab}[1]{#1}
\providecommand{\url}[1]{\texttt{#1}}
\expandafter\ifx\csname urlstyle\endcsname\relax
  \providecommand{\doi}[1]{doi: #1}\else
  \providecommand{\doi}{doi: \begingroup \urlstyle{rm}\Url}\fi

\bibitem[{Airbus Defence and
  Space}(2013)]{AirbusDefenceandSpace2013SPOTIntelligence}
{Airbus Defence and Space}.
\newblock {SPOT 6 | SPOT 7 High Resolution Broad Coverage Intelligence}.
\newblock Technical report, 2013.
\newblock URL \url{www.geostore.com}.

\bibitem[Aitken et~al.(2017)Aitken, Ledig, Theis, Caballero, Wang, and
  Shi]{Aitken2017CheckerboardResize}
Andrew Aitken, Christian Ledig, Lucas Theis, Jose Caballero, Zehan Wang, and
  Wenzhe Shi.
\newblock {Checkerboard artifact free sub-pixel convolution: A note on
  sub-pixel convolution, resize convolution and convolution resize}.
\newblock 7 2017.
\newblock URL \url{http://arxiv.org/abs/1707.02937}.

\bibitem[Arnab et~al.(2018)Arnab, Zheng, Jayasumana, Romera-Paredes, Larsson,
  Kirillov, Savchynskyy, Rother, Kahl, and
  Torr]{Arnab2018ConditionalPrediction}
Anurag Arnab, Shuai Zheng, Sadeep Jayasumana, Bernardino Romera-Paredes, Måns
  Larsson, Alexander Kirillov, Bogdan Savchynskyy, Carsten Rother, Fredrik
  Kahl, and Philip~H.S. Torr.
\newblock {Conditional Random Fields Meet Deep Neural Networks for Semantic
  Segmentation: Combining Probabilistic Graphical Models with Deep Learning for
  Structured Prediction}.
\newblock \emph{IEEE Signal Processing Magazine}, 35\penalty0 (1):\penalty0
  37--52, 1 2018.
\newblock ISSN 10535888.
\newblock \doi{10.1109/MSP.2017.2762355}.

\bibitem[Artacho and Savakis(2019)]{Artacho2019WaterfallSegmentation}
Bruno Artacho and Andreas Savakis.
\newblock {Waterfall Atrous Spatial Pooling Architecture for Efficient Semantic
  Segmentation}.
\newblock \emph{Sensors}, 19\penalty0 (24):\penalty0 5361, 12 2019.
\newblock ISSN 1424-8220.
\newblock \doi{10.3390/s19245361}.
\newblock URL \url{https://www.mdpi.com/1424-8220/19/24/5361}.

\bibitem[Bochkovskiy et~al.(2020)Bochkovskiy, Wang, and
  Liao]{Bochkovskiy2020YOLOv4:Detection}
Alexey Bochkovskiy, Chien-Yao Wang, and Hong-Yuan~Mark Liao.
\newblock {YOLOv4: Optimal Speed and Accuracy of Object Detection}.
\newblock 4 2020.
\newblock URL \url{http://arxiv.org/abs/2004.10934}.

\bibitem[Campbell and Wynne(2011)]{Campbell2011IntroductionEdition}
J~B Campbell and R~H Wynne.
\newblock \emph{{Introduction to Remote Sensing, Fifth Edition}}.
\newblock Guilford Publications, 2011.
\newblock ISBN 9781609181772.
\newblock URL \url{https://books.google.nl/books?id=NkLmDjSS8TsC}.

\bibitem[Chen et~al.(2018{\natexlab{a}})Chen, Papandreou, Kokkinos, Murphy, and
  Yuille]{Chen2018DeepLab:CRFs}
Liang~Chieh Chen, George Papandreou, Iasonas Kokkinos, Kevin Murphy, and
  Alan~L. Yuille.
\newblock {DeepLab: Semantic Image Segmentation with Deep Convolutional Nets,
  Atrous Convolution, and Fully Connected CRFs}.
\newblock \emph{IEEE Transactions on Pattern Analysis and Machine
  Intelligence}, 40\penalty0 (4):\penalty0 834--848, 4 2018{\natexlab{a}}.
\newblock ISSN 01628828.
\newblock \doi{10.1109/TPAMI.2017.2699184}.
\newblock URL \url{http://liangchiehchen.com/projects/}.

\bibitem[Chen et~al.(2018{\natexlab{b}})Chen, Zhu, Papandreou, Schroff, and
  Adam]{Chen2018Encoder-decoderSegmentation}
Liang~Chieh Chen, Yukun Zhu, George Papandreou, Florian Schroff, and Hartwig
  Adam.
\newblock {Encoder-decoder with atrous separable convolution for semantic image
  segmentation}.
\newblock In \emph{Lecture Notes in Computer Science (including subseries
  Lecture Notes in Artificial Intelligence and Lecture Notes in
  Bioinformatics)}, volume 11211 LNCS, pages 833--851. Springer Verlag,
  2018{\natexlab{b}}.
\newblock ISBN 9783030012335.
\newblock \doi{10.1007/978-3-030-01234-2{\_}49}.

\bibitem[Cheng and Han(2016)]{Cheng2016AImages}
Gong Cheng and Junwei Han.
\newblock {A survey on object detection in optical remote sensing images}.
\newblock \emph{ISPRS Journal of Photogrammetry and Remote Sensing},
  117:\penalty0 11--28, 7 2016.
\newblock \doi{10.1016/j.isprsjprs.2016.03.014}.
\newblock URL \url{https://doi.org/10.1016/j.isprsjprs.2016.03.014}.

\bibitem[Dai et~al.(2017)Dai, Qi, Xiong, Li, Zhang, Hu, and
  Wei]{Dai2017DeformableNetworks}
Jifeng Dai, Haozhi Qi, Yuwen Xiong, Yi~Li, Guodong Zhang, Han Hu, and Yichen
  Wei.
\newblock {Deformable Convolutional Networks}.
\newblock In \emph{Proceedings of the IEEE International Conference on Computer
  Vision}, volume 2017-October, pages 764--773. Institute of Electrical and
  Electronics Engineers Inc., 12 2017.
\newblock ISBN 9781538610329.
\newblock \doi{10.1109/ICCV.2017.89}.
\newblock URL \url{https://github.com/}.

\bibitem[Green and Sussman(1990)]{Green1990DeforestationImages}
Glen~M Green and Robert~W Sussman.
\newblock {Deforestation history of the eastern rain forests of Madagascar from
  satellite images}.
\newblock \emph{Science}, 248\penalty0 (4952):\penalty0 212--215, 1990.

\bibitem[He et~al.(2014)He, Zhang, Ren, and Sun]{He2014SpatialRecognition}
Kaiming He, Xiangyu Zhang, Shaoqing Ren, and Jian Sun.
\newblock {Spatial Pyramid Pooling in Deep Convolutional Networks for Visual
  Recognition}.
\newblock \emph{Lecture Notes in Computer Science (including subseries Lecture
  Notes in Artificial Intelligence and Lecture Notes in Bioinformatics)}, 8691
  LNCS\penalty0 (PART 3):\penalty0 346--361, 6 2014.
\newblock \doi{10.1007/978-3-319-10578-9{\_}23}.
\newblock URL \url{http://arxiv.org/abs/1406.4729
  http://dx.doi.org/10.1007/978-3-319-10578-9_23}.

\bibitem[Howard and Gugger(2020)]{Howard2020Fastai:Learning}
Jeremy Howard and Sylvain Gugger.
\newblock {fastai: A Layered API for Deep Learning}.
\newblock \emph{Information (Switzerland)}, 11\penalty0 (2), 2 2020.
\newblock \doi{10.3390/info11020108}.
\newblock URL \url{http://arxiv.org/abs/2002.04688
  http://dx.doi.org/10.3390/info11020108}.

\bibitem[{Humboldt State
  University}(2019)]{HumboldtStateUniversity2019GSPResolution}
{Humboldt State University}.
\newblock {GSP 216 Introduction to remote sensing, Humboldt State University:
  Resolution}.
\newblock
  http://gsp.humboldt.edu/OLM/Courses/GSP{\_}216{\_}Online/lesson3-1/resolution.html,
  2019.
\newblock URL
  \url{http://gsp.humboldt.edu/OLM/Courses/GSP_216_Online/lesson3-1/resolution.html}.

\bibitem[Ko(2018)]{Ko2018USAMaintenance}
Rita Ko.
\newblock {USA for UNHCR Launches Satellite Imagery and Crowdsourcing Project
  to Improve Refugee Camp Planning and Maintenance}.
\newblock
  https://www.unrefugees.org/news/usa-for-unhcr-launches-satellite-imagery-and-crowdsourcing-project-to-improve-refugee-camp-planning-and-maintenance/,
  2018.
\newblock URL \url{https://bit.ly/3kW2MEU}.

\bibitem[Lin et~al.(2017)Lin, Goyal, Girshick, He, and
  Doll{\'{a}}r]{Lin2017FocalDetection}
Tsung-Yi Lin, Priya Goyal, Ross Girshick, Kaiming He, and Piotr Doll{\'{a}}r.
\newblock {Focal Loss for Dense Object Detection}.
\newblock 8 2017.
\newblock URL \url{http://arxiv.org/abs/1708.02002}.

\bibitem[Liu et~al.(2019)Liu, Wang, and Cheng]{Liu2019AnDeeplabv3+}
Jiaqi Liu, Zhili Wang, and Kangxin Cheng.
\newblock {An Improved Algorithm for Semantic Segmentation of Remote Sensing
  Images Based on Deeplabv3+}.
\newblock In \emph{Proceedings of the 5th International Conference on
  Communication and Information Processing}, New York, NY, USA, 2019. ACM.
\newblock ISBN 9781450372589.
\newblock URL \url{https://doi.org/10.1145/3369985.3370027}.

\bibitem[Long et~al.(2014)Long, Shelhamer, and
  Darrell]{Long2014FullySegmentation}
Jonathan Long, Evan Shelhamer, and Trevor Darrell.
\newblock {Fully Convolutional Networks for Semantic Segmentation}.
\newblock \emph{IEEE Transactions on Pattern Analysis and Machine
  Intelligence}, 39\penalty0 (4):\penalty0 640--651, 11 2014.
\newblock URL \url{http://arxiv.org/abs/1411.4038}.

\bibitem[Majid~Azimi et~al.(2019)Majid~Azimi, Henry, Sommer, Schumann, and
  Vig]{MajidAzimi2019SkyScapes-Fine-GrainedScenes}
Seyed Majid~Azimi, Corentin Henry, Lars Sommer, Arne Schumann, and Eleonora
  Vig.
\newblock {SkyScapes-Fine-Grained Semantic Understanding of Aerial Scenes}.
\newblock Technical report, 2019.
\newblock URL
  \url{https://www.dlr.de/eoc/en/desktopdefault.aspx/tabid-12760Aerialimagewithoverlaidannotation:dense}.

\bibitem[Parmar et~al.(2020)Parmar, Bhatia, Negi, and
  Suri]{Parmar2020ExplorationDrones}
Vivek Parmar, Narayani Bhatia, Shubham Negi, and Manan Suri.
\newblock {Exploration of Optimized Semantic Segmentation Architectures for
  edge-Deployment on Drones}.
\newblock 7 2020.
\newblock URL \url{http://arxiv.org/abs/2007.02839}.

\bibitem[Pilkington et~al.(2019)Pilkington, Svetlichnaya, and
  Holmes]{Pilkington2019GitHubBenchmark}
Nicholas Pilkington, Stacey Svetlichnaya, and Tom Holmes.
\newblock {GitHub - dronedeploy/dd-ml-segmentation-benchmark: DroneDeploy
  Machine Learning Segmentation Benchmark}, 2019.
\newblock URL
  \url{https://github.com/dronedeploy/dd-ml-segmentation-benchmark}.

\bibitem[Richards and Friess(2016)]{Richards2016Rates20002012}
Daniel~R Richards and Daniel~A Friess.
\newblock {Rates and drivers of mangrove deforestation in Southeast Asia,
  2000–2012}.
\newblock \emph{Proceedings of the National Academy of Sciences}, 113\penalty0
  (2):\penalty0 344--349, 2016.

\bibitem[Ronneberger et~al.(2015)Ronneberger, Fischer, and
  Brox]{Ronneberger2015U-net:Segmentation}
Olaf Ronneberger, Philipp Fischer, and Thomas Brox.
\newblock {U-net: Convolutional networks for biomedical image segmentation}.
\newblock In \emph{Lecture Notes in Computer Science (including subseries
  Lecture Notes in Artificial Intelligence and Lecture Notes in
  Bioinformatics)}, volume 9351, pages 234--241. Springer Verlag, 2015.
\newblock ISBN 9783319245737.
\newblock \doi{10.1007/978-3-319-24574-4{\_}28}.

\bibitem[Roser and Ortiz-Ospina(2013)]{Roser2013GlobalPoverty}
Max Roser and Esteban Ortiz-Ospina.
\newblock {Global Extreme Poverty}.
\newblock \emph{Our World in Data}, 2013.

\bibitem[Sang and Minh(2018)]{Sang2018FullySegmentation}
Dinh~Viet Sang and Nguyen~Duc Minh.
\newblock {Fully residual convolutional neural networks for aerial image
  segmentation}.
\newblock In \emph{ACM International Conference Proceeding Series}, pages
  289--296, New York, New York, USA, 12 2018. Association for Computing
  Machinery.
\newblock ISBN 9781450365390.
\newblock \doi{10.1145/3287921.3287970}.
\newblock URL \url{http://dl.acm.org/citation.cfm?doid=3287921.3287970}.

\bibitem[Shermeyer et~al.(2020)Shermeyer, Hogan, Brown, Van~Etten, Weir,
  Pacifici, Haensch, Bastidas, Soenen, Bacastow, and
  Lewis]{Shermeyer2020SpaceNetDataset}
Jacob Shermeyer, Daniel Hogan, Jason Brown, Adam Van~Etten, Nicholas Weir,
  Fabio Pacifici, Ronny Haensch, Alexei Bastidas, Scott Soenen, Todd Bacastow,
  and Ryan Lewis.
\newblock {SpaceNet 6: Multi-Sensor All Weather Mapping Dataset}.
\newblock pages 768--777, 4 2020.
\newblock URL \url{http://arxiv.org/abs/2004.06500}.

\bibitem[Soman(2019)]{Soman2019RooftopImagery}
Kritik Soman.
\newblock {Rooftop detection using aerial drone imagery}.
\newblock In \emph{ACM International Conference Proceeding Series}, pages
  281--284, New York, New York, USA, 1 2019. Association for Computing
  Machinery.
\newblock ISBN 9781450362078.
\newblock \doi{10.1145/3297001.3297041}.
\newblock URL \url{http://dl.acm.org/citation.cfm?doid=3297001.3297041}.

\bibitem[Tao et~al.(2020)Tao, Sapra, and
  Catanzaro]{Tao2020HierarchicalSegmentation}
Andrew Tao, Karan Sapra, and Bryan Catanzaro.
\newblock {Hierarchical Multi-Scale Attention for Semantic Segmentation}.
\newblock 5 2020.
\newblock URL \url{http://arxiv.org/abs/2005.10821}.

\bibitem[Viet~Sang and Duc~Minh(2018)]{VietSang2018FullySegmentation}
Dinh Viet~Sang and Nguyen Duc~Minh.
\newblock {Fully Residual Con-volutional Neural Networks for, Aerial Image
  Segmentation}.
\newblock In \emph{Proceedings of the Ninth International Symposium on
  Information and Communication Technology - SoICT 2018}, New York, New York,
  USA, 2018. ACM Press.
\newblock ISBN 9781450365390.
\newblock URL \url{https://doi.org/10.1145/3287921.3287970}.

\bibitem[Wang et~al.(2019)Wang, Zhang, and Wu]{Wang2019AccurateImage}
Shuqi Wang, Chuang Zhang, and Ming Wu.
\newblock {Accurate Semantic Segmentation in Remote Sensing Image}.
\newblock In \emph{Proceedings of the 2019 8th International Conference on
  Computing and Pattern Recognition}, pages 173--178, New York, NY, USA, 10
  2019. ACM.
\newblock ISBN 9781450376570.
\newblock \doi{10.1145/3373509.3373535}.
\newblock URL \url{https://dl.acm.org/doi/10.1145/3373509.3373535}.

\bibitem[Wurm et~al.(2019)Wurm, Stark, Zhu, Weigand, and
  Taubenb{\"{o}}ck]{Wurm2019SemanticNetworks}
Michael Wurm, Thomas Stark, Xiao~Xiang Zhu, Matthias Weigand, and Hannes
  Taubenb{\"{o}}ck.
\newblock {Semantic segmentation of slums in satellite images using transfer
  learning on fully convolutional neural networks}.
\newblock \emph{ISPRS Journal of Photogrammetry and Remote Sensing},
  150:\penalty0 59--69, 4 2019.
\newblock ISSN 09242716.
\newblock \doi{10.1016/j.isprsjprs.2019.02.006}.

\bibitem[Zhou(2003)]{Zhou2003FutureSatellites}
Guoqing Zhou.
\newblock {Future Intelligent Earth Observing Satellites}.
\newblock \emph{Proc SPIE}, 5151, 1 2003.
\newblock \doi{10.1117/12.501232}.

\bibitem[Zhu et~al.(2018)Zhu, Sapra, Reda, Shih, Newsam, Tao, and
  Catanzaro]{Zhu2018ImprovingRelaxation}
Yi~Zhu, Karan Sapra, Fitsum~A. Reda, Kevin~J. Shih, Shawn Newsam, Andrew Tao,
  and Bryan Catanzaro.
\newblock {Improving Semantic Segmentation via Video Propagation and Label
  Relaxation}.
\newblock \emph{Proceedings of the IEEE Computer Society Conference on Computer
  Vision and Pattern Recognition}, 2019-June:\penalty0 8848--8857, 12 2018.
\newblock URL \url{http://arxiv.org/abs/1812.01593}.

\bibitem[Zoph et~al.(2020)Zoph, Ghiasi, Lin, Cui, Liu, Cubuk, and
  Le]{Zoph2020RethinkingSelf-training}
Barret Zoph, Golnaz Ghiasi, Tsung-Yi Lin, Yin Cui, Hanxiao Liu, Ekin~D. Cubuk,
  and Quoc~V. Le.
\newblock {Rethinking Pre-training and Self-training}.
\newblock 6 2020.
\newblock URL \url{http://arxiv.org/abs/2006.06882}.

\end{thebibliography}

\end{document}